\pdfoutput=1

\documentclass[11pt]{article}

\usepackage{acl}

\usepackage{times}
\usepackage{latexsym}
\usepackage[T1]{fontenc}
\usepackage[utf8]{inputenc}
\usepackage{microtype}

\usepackage{hyperref}
\usepackage{url}
\usepackage{booktabs}
\usepackage{amsfonts}
\usepackage{graphicx}
\usepackage{listings}
\usepackage[normalem]{ulem}
\usepackage{multirow}
\usepackage{array}
\usepackage{makecell}
\usepackage{tabularx}
\usepackage{geometry}

\usepackage{ulem}

\newcommand{\cblu}{\color{blue}}
\newcommand{\cred}{\color{red}}

\long\def\BOC#1\EOC{\message{(Commented text )}}
\long\def\BOCC#1\EOCC{\message{(Commented text )}}
\long\def\BOCCC#1\EOCCC{\message{(Commented text )}}

\long\def\NBB#1{}

\usepackage{tcolorbox}
\tcbuselibrary{theorems,listings,skins,breakable,minted}

\def\ar{\leftarrow}
\def\bi{\begin{itemize}}
\def\ei{\end{itemize}}
\def\beq{\begin{equation}}
\def\eeq#1{\label{#1}\end{equation}}
\def\ba{\begin{array}}
\def\ea{\end{array}}
\def\i#1{\hbox{\it #1\/}}

\def\no{\i{not}}

\def\ar{\leftarrow}

\def\i#1{\hbox{\itshape #1\/}}

\title{LLMs as ASP Programmers: Self-Correction Enables \\ Task-Agnostic Nonmonotonic Reasoning}

\author{Adam Ishay $^1$\\
  $^1$ Arizona State University \\
  \texttt{aishay@asu.edu} \\\And
  Joohyung Lee $^{1,2}$ \thanks{Corresponding Author}\\
  $^2$ Samsung Research \\
  \texttt{joolee@asu.edu} \\}

\begin{document}

\maketitle   

\begin{abstract}
Recent large language models (LLMs) have achieved impressive reasoning milestones but continue to struggle with high computational costs, logical inconsistencies, and sharp performance degradation on high-complexity problems. While neuro-symbolic methods attempt to mitigate these issues by coupling LLMs with symbolic reasoners, existing approaches typically rely on monotonic logics (e.g., SMT) that cannot represent defeasible reasoning—essential components of human cognition. We present {\sf LLM+ASP}, a framework that translates natural language into Answer Set Programming (ASP), a nonmonotonic formalism based on stable model semantics. Unlike prior {\sf LLM+ASP} approaches that require manually authored knowledge modules, domain-specific prompts, or evaluation restricted to single problem classes, our framework operates without any per-task engineering and applies uniformly across diverse reasoning tasks. Our system utilizes an automated self-correction loop where structured feedback from the ASP solver enables iterative refinement. Evaluating across six diverse benchmarks, we demonstrate that: (1) stable model semantics allow LLMs to naturally express default rules and exceptions, outperforming SMT-based alternatives by significant margins on nonmonotonic tasks; (2) iterative self-correction is the primary driver of performance, effectively replacing the need for handcrafted domain knowledge; (3) compact in-context reference guides substantially outperform verbose documentation, revealing a ``context rot" phenomenon where excessive context hinders constraint adherence.
\end{abstract}
%

\section{Introduction} \label{sec:intro} 

Recent advancements in Large Language Models (LLMs) have pushed the boundaries of automated reasoning. Models like Google's Gemini series, OpenAI's o-series, and DeepSeek's R1 \cite{gemini2_5thinking25, jaech24openai, o3o4mini24, guo25deepseek} are often referred to as ``Reasoning" models. Unlike previous LLMs, these models solve complex, multi-step problems by generating explicit intermediate ``thinking" tokens, reflecting on them, and self-correcting their reasoning.

By scaling computation during inference, these models achieve impressive performance on many complex reasoning tasks. However, they incur high computational costs, and their performance falls sharply as problems surpass a certain complexity threshold \cite{shojaee25illusion,chen24not, sui25stop}. Furthermore, despite generating long reasoning traces, they often exhibit logical inconsistencies between their intermediate ``thinking" tokens and final answers \cite{chen25reasoning, stechly25beyond}.

A promising path to mitigate these weaknesses lies in neuro-symbolic approaches that combine the natural language strengths of LLMs with the rigor of external symbolic reasoners. While promising, these methods have often been restricted to narrow domains and require careful prompting or human intervention, hindering their scalability and general applicability \cite{lin24clmasp, kambhampati24position, guan23leveraging}.

A crucial limitation of many existing neuro-symbolic approaches is their reliance on monotonic logics (e.g., First-Order Logic, Satisfiability Modulo Theories).
In these frameworks, once a conclusion is derived, it cannot be retracted with the addition of new information, making them ill-suited for representing defeasible reasoning, preferences, or exceptions. In contrast, human reasoning is inherently nonmonotonic: we routinely make default assumptions (e.g., "birds fly") that we retract upon encountering specific exceptions (e.g., "penguins").

Our work introduces {\sf LLM+ASP}, which bridges this gap by coupling LLMs with Answer Set Programming (ASP) \cite{lif08,brewka11answer,lee08a,gll06}. ASP’s theoretical foundation in stable model semantics treats default negation and alternative possible worlds as first-class constructs, allowing LLMs to model complex logical conflicts more naturally than in FOL or SMT.

\noindent
{\bf Distinction from prior LLM+ASP work.} Our earlier work explored coupling LLMs with ASP but relied on task-specific engineering. \citet{yang23coupling} generated partial ASP representations requiring manually authored knowledge modules and few-shot examples per problem class; \citet{ishay23leveraging} generated complete programs via custom-designed prompting but evaluated only on logic grid puzzles. Critically, both treated the LLM as a one-shot translator with no mechanism to recover from generation errors—the LLM produced an ASP program, and if it was flawed, there was no path to correction. In contrast, LLM+ASP addresses this by introducing an automated self-correction loop driven by solver feedback, requiring no per-task prompts, no few-shot examples, and no manually authored domain knowledge. This enables a single task-agnostic framework to operate across six diverse benchmarks spanning constraint satisfaction, planning, and nonmonotonic reasoning—substantially broader than any prior LLM+ASP evaluation.

\noindent{\bf Enabling this task-agnostic approach is a key empirical observation.}
Recent LLMs, especially those scaled up for test-time computation, can generate ASP code with relatively little guidance, likely due to sufficient ASP code exposure in their training data. However, an LLM's initial attempt at writing an ASP program is often flawed. To remedy this, we introduce {\sf LLM+ASP}, a framework that enables LLMs to become effective ASP programmers through an automated feedback loop. While the core principle of using execution feedback for iterative refinement builds on established work in semantic parsing and program synthesis~\citep{chen23teaching,olausson23self}, our contribution lies in demonstrating that this approach, when applied to ASP with its nonmonotonic semantics, enables robust automated reasoning across diverse task types without task-specific engineering. Specifically, we show that: (1) ASP's stable model semantics provides a more natural target for LLM-based code generation in nonmonotonic reasoning tasks compared to monotonic alternatives, (2) a compact in-context reference guide substantially outperforms verbose documentation, revealing a ``context rot" phenomenon, and (3) the self-correction loop is the primary performance driver, with reference materials playing a secondary but important role.

We find that recent Reasoning models are particularly adept at this process, generating correct ASP programs by reflecting on solver output messages. Here, thinking tokens are used to generate and revise ASP programs that guide efficient symbolic search, rather than directly traversing the search space. Additionally, supplying LLMs with reference materials on ASP improves performance, especially for weaker models.

Our code implementation is available on GitHub: \url{https://github.com/azreasoners/llm-asp-programmer}.

\section{Related Work} \label{sec:preliminaries} 

\subsection{Reasoning with LLMs}

Recent large reasoning models—Google's Gemini series~\citep{gemini2_5thinking25}, OpenAI's o-series~\citep{jaech24openai,o3o4mini24}, and DeepSeek's R1~\citep{guo25deepseek}—achieve impressive performance by scaling test-time computation. However, they face three key limitations: high computational costs~\citep{chen24not,sui25stop}, sharp performance degradation beyond complexity thresholds~\citep{shojaee25illusion,valmeekam25systematic}, and logical inconsistencies between reasoning traces and answers~\citep{chen25reasoning,stechly25beyond}.

\paragraph{Neuro-Symbolic Alternatives.}
Coupling LLMs with symbolic reasoners consistently outperforms direct LLM planning~\citep{kalyanpur24llm,liu23llm+,pan23logic,hao25large}. However, existing approaches require manually authored domain knowledge~\citep{liu23llm+,xie23translating}, hand-designed schemas~\citep{kalyanpur24llm}, human-in-the-loop intervention~\citep{guan23leveraging}, or task-specific prompt engineering~\citep{pan23logic,hao25large}.

It has been shown that LLMs have the capacity to improve their performance by self-correction via various forms of feedback~\citep{madaan23self,shinn23reflexion,kamoi24llms,stechly24self}, and program synthesis systems use compiler feedback for iterative refinement~\citep{chen23teaching,olausson23self}. The LLM-Modulo framework~\citep{kambhampati24position} provides structured feedback but requires correct initial models.

\smallskip\noindent
{\bf LLMs with Answer Set Programming.}\ \ 
%
%
Prior LLM+ASP approaches \cite{yang23coupling,ishay23leveraging} treated the LLM as a one-shot translator requiring task-specific engineering, without mechanisms to recover from generation errors, as detailed in Section~\ref{sec:intro}. 
Alternatively, \citet{ishay25llm+} explored using action language ${\cal BC}$+ \cite{babb15action,babb15action1} in place of ASP with self-revisions, but focused primarily on  planning problems. 
More broadly, other neuro-symbolic approaches coupling LLMs with symbolic reasoners similarly require manual expert knowledge or restriction to narrow domains \cite{lin24clmasp,kalyanpur24llm}. Our automated self-correction loop, which we identify as the primary performance driver (Section~\ref{ssec:rq3}), addresses this shared limitation across existing work.

\subsection{Answer Set Programming} \label{ssec:asp-prelim} 
Answer Set Programming (ASP) \cite{lif08,brewka11answer} is a knowledge representation formalism based on stable model semantics \cite{gel88}, capable of expressing causal reasoning, default reasoning, aggregates, and constraints. 

ASP's declarative nature allows users to specify \emph{what} must be true rather than \emph{how} to find solutions, making it ideal for problems defined by complex logical constraints. Efficient solvers like {\sc clingo} use Conflict-Driven Clause Learning to handle NP-hard problems and beyond.

ASP rules of the form $A \ar B, \no\ C$, where $\no$ is default negation, read as ``derive $A$ if one can derive $B$ and cannot derive $C$." For example, \texttt{flies(X) $\ar$ bird(X), not penguin(X)} means $X$ flies if $X$ is a bird unless there is evidence it is a penguin. Facts like \texttt{bird(tweety)} have no body. Constraints like  \texttt{$\ar$ flies(X), penguin(X)} forbid patterns. Modern ASP includes aggregates (\texttt{\#count}, \texttt{\#sum}), choice rules (e.g., \texttt{\{color(X,red); color(X,blue)\}}), and optimization (\texttt{\#minimize}).
For more on {\sc clingo}, see \cite{lifschitz19answer} or the user guide.\footnote{\url{https://github.com/potassco/guide/releases}.}

While ASP is useful for automated reasoning, writing ASP code typically requires expertise. Prior work required carefully designed prompts, manual knowledge modules, or human intervention. We demonstrate that recent LLMs, integrated into a self-correction loop, enable automated ASP generation without human involvement.

\section{Our Method} \label{sec:method} 

\begin{figure*}
    \centering
    \includegraphics[width=0.9\linewidth,height=5.1cm]{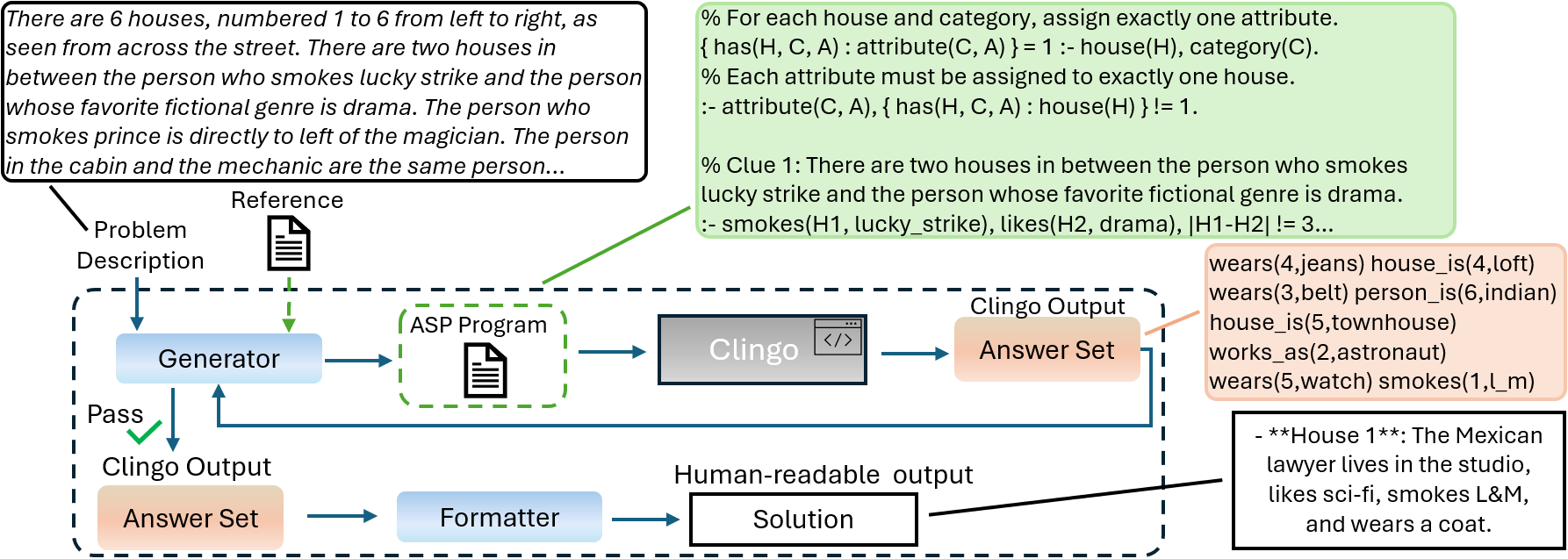}
    \caption{{\sf LLM+ASP} Pipeline}
    \label{fig:main-figure}
    \vspace{-1em}
\end{figure*}
 
Our approach leverages LLMs to encode natural language knowledge into ASP for efficient reasoning.

The {\sf LLM+ASP} pipeline (Figure~\ref{fig:main-figure}) begins with an LLM \textit{generator} receiving a natural language problem description. The generator writes or updates the symbolic ASP representation using a short instruction prompt (Appendix~\ref{appendix:prompts}), which contains the problem description, task-agnostic instructions, and an optional ASP reference guide for in-context learning. Here, \emph{task-agnostic} means that the prompt contains no task-specific prompt engineering: it uses no per-task prompts, no few-shot task examples, and no manually authored domain knowledge.  
We experiment with two references: 
\textbf{(i)} \textit{standard reference}: Gemini 2.5 Pro summary of the official {\sc clingo} manual (22k tokens). 
\textbf{(ii)} \textit{compact reference}: derived from (i), restricted to common ASP constructs, such as basic rule syntax, variable conventions, common operators, pooling, and standard constraint formats (2.6k tokens; one-eighth the size).\footnote{Appendix~\ref{appendix:compact-reference} presents the compact reference as well as the method used to generate it.} 
While reference creation requires one-time manual effort, it eliminates task-specific prompt engineering for each problem, unlike prior approaches requiring extensive per-domain customization~\citep{hao25large,pan23logic}.

Once generated, the program is executed by {\sc clingo} (we used v5.8.0). The output is fed back to the generator, which decides whether to accept the proposed answer set (i.e., stable model) or update the ASP code based on solver feedback. The generator is given instructions to either ``UPDATE" or ``PASS." 
An update rewrites the program based on {\sc clingo} feedback. A pass is chosen only when the LLM judges the output correct. If passing, the program and solution are accepted. Otherwise, the cycle repeats until a solution passes or maximum revisions are reached.
An example trace is in Appendix~\ref{appendix:example}. 

\section{Experiments} \label{sec:experiments} 

To evaluate the effectiveness and characteristics of our {\sf LLM+ASP} framework, we conduct a series of experiments designed to answer the following key research questions:

\begin{itemize} 
\setlength{\itemsep}{0pt}
\setlength{\parskip}{0pt}
\item[RQ1] {\bf Overall Performance:}  
How does the {\sf LLM+ASP} framework perform on complex reasoning tasks compared to standalone LLMs?
Does the framework benefit LLMs of varying strengths differently? 
\item[RQ2] {\bf Impact of In-Context Knowledge:}  
How does providing an external reference on ASP syntax and conventions affect the performance of LLMs with different levels of pre-existing knowledge? 
\item[RQ3] {\bf Impact of Iterative Correction: }
What is the effect of self-revision on program quality and reasoning performance? How does model capability affect the revision process? What are the primary failure cases of {\sf LLM+ASP}?
\item[RQ4] {\bf Preliminary Comparison with SMT:}
If we replace ASP with SMT in our pipeline, how would it perform, particularly on nonmonotonic reasoning tasks where ASP has theoretical advantages?
\end{itemize}

\begin{table*}[t]
\centering
\fontsize{8}{9}\selectfont 
\caption{Main Benchmark Results}\vspace{-1em}
\label{tab:main-result}
\setlength{\tabcolsep}{3.5pt}
\begin{tabular}{l|c|c|c|c|c|c|c|c}
\toprule
\multirow{2}{*}{\textbf{Model}} & \multirow{1}{*}{\textbf{ZL-XL}} & \multicolumn{1}{c|}{\textbf{ZL-XXL (\%)}} & \multicolumn{1}{c|}{\textbf{SudokuBench }} & \multirow{1}{*}{\textbf{MysteryBW}} & \multicolumn{2}{c|}{\textbf{MultiLogicNMR (\%)}}{} & \multirow{1}{*}{\textbf{Board-}} & \multirow{1}{*}{\textbf{Average}} \\
\cmidrule(lr){6-7}
& (\%) & (\%) & (\%) & (\%) & Skeptical & Credulous & \textbf{gameQA} (\%) & (\%) \\
\midrule
\multicolumn{9}{c}{\textbf{Baselines}} \\
\midrule
Gemini 2.5 Pro    & $74.7 \pm 0.47$ & $32.3 \pm 1.70$ & $10.3 \pm 0.47$ & $\textbf{83.7} \pm 3.30$ & $\textbf{88.5} \pm 2.16$ & $61.0 \pm 2.55$ & $89.8 \pm 1.25$ & 62.9 \\
Gemini 2.5 Flash  & $38.0 \pm 2.16$ & $20.0 \pm 0.82$ & $7.3 \pm 0.47$ & $66.3 \pm 2.62$ & $42.3 \pm 0.47$ & $36.2 \pm 0.62$ & $\textbf{93.0} \pm 0.71$ & 43.3 \\
o4-mini           & $\textbf{88.7} \pm 2.05$ & $\textbf{34.0} \pm 1.41$ & $\textbf{15.0} \pm 2.94$ & $81.0 \pm 2.45$ & $67.5 \pm 1.78$ & $\textbf{69.7} \pm 0.24$ & $88.2 \pm 1.55$ & \textbf{63.4} \\
DS-R1-0528  & $83.0 \pm 3.27$ & $27.7 \pm 0.47$ & $8.3 \pm 1.70$ & $79.7 \pm 6.94$ & $39.3 \pm 0.62$ & $46.8 \pm 1.31$ & $81.0 \pm 1.87$ & 52.3 \\
DS-V3 & $42.0 \pm 3.56$ & $7.0 \pm 1.63$ & $1.0 \pm 0.00$ & $10.3 \pm 1.70$ & $57.7 \pm 1.65$ & $51.0 \pm 3.27$ & $71.3 \pm 0.85$ & 34.3 \\
\midrule
 Average &  $65.3$ & $24.2$ & $8.4$ & $64.2$ & $59.0$ & $52.9$ & $84.7$ & $51.3$ \\ 
\midrule
\multicolumn{9}{c}{\textbf{LLM + ASP using Compact Reference}} \\
\midrule
Gem. 2.5 Pro+ASP    & $ 96.7 \pm 0.47$ & $\textbf{97.7} \pm 1.25$ & $\textbf{74.7} \pm 1.25$ & $\textbf{98.3} \pm 0.47$ & $\textbf{100.0} \pm 0.00$ & $93.8 \pm 1.03$ & $\textbf{96.3} \pm 0.47$ & \textbf{93.9} \\
Gem. 2.5 Flash+ASP  & $84.3 \pm 1.25$ & $70.3 \pm 1.25$ & $65.7 \pm 2.05$ & $96.7 \pm 0.94$ & $ 98.3 \pm 0.47$ & $\textbf{98.3} \pm 0.47$ & $94.7 \pm 0.85$ & 86.9 \\
o4-mini+ASP           & $95.0 \pm 0.82$ & $76.7 \pm 2.49$ & $67.0 \pm 1.41$ & $72.7 \pm 2.62$ & $95.8 \pm 1.65$ & $84.0 \pm 3.08$ & $85.2 \pm 0.47$ & 82.3 \\
DS-R1-0528+ASP  & $\textbf{97.3} \pm 0.47$ & $75.0 \pm 1.63$ & $70.0 \pm 1.63$ & $81.7 \pm 2.49$ & $83.8 \pm 0.85$ & $73.5 \pm 1.87$ & $91.7 \pm 1.55$ & 81.9 \\
DS-V3+ASP       & $69.0 \pm 5.35$ & $37.0 \pm 5.10$ & $21.7 \pm 2.87$ & $13.7 \pm 1.70$ & $74.3 \pm 1.84$ & $55.3 \pm 1.25$ & $44.0 \pm 2.94$ & 45.0 \\
\midrule
Average &  $88.5$ & $71.3$ & $59.8$ & $72.6$ & $90.5$ & $81.0$ & $82.4$ & $78.0$ \\ 
\bottomrule
\end{tabular}
\end{table*}

\noindent
\textbf{Benchmarks.} 
We selected a diverse suite of benchmarks to rigorously evaluate the {\sf LLM+ASP} framework across a wide spectrum of logical reasoning challenges.

\textbf{ZebraLogic} \cite{richardson25zebra} evaluates logical reasoning using grid puzzles. From this benchmark, we use 100 of the hardest ``XL" puzzles (\textbf{ZL-XL}) and created a custom-built set of even more challenging problems, which we named \textbf{ZL-XXL}, designed to enable finer-grained evaluation. The dataset is categorized into five sizes: $n \times n$ for $n \in \{4, 6, 8, 10, 12\}$, with 20 puzzles in each category. 
\textbf{SudokuBench} \cite{seely25sudoku} is a curated collection of 100 challenging and unconventional Sudoku variants. Unlike standard Sudoku, its puzzles introduce novel, interacting logical constraints. 
We include 100 problems from the \textbf{Mystery Blocksworld} benchmark \cite{valmeekam23planbench}, a planning domain variant, intentionally obfuscated to require models to reason through transformations.

 To explicitly evaluate nonmonotonic reasoning, we include two benchmarks that go beyond monotonic logical inference. \textbf{MultiLogicNMR} \cite{xiu25multilogicnmr} is a benchmark for nonmonotonic reasoning, covering default reasoning, exceptions, and rule conflicts. This allows us to directly assess a model's ability to handle inference retraction and alternative possible conclusions. Specifically, the benchmark distinguishes between \textit{skeptical} reasoning, in which a conclusion must be entailed by all models, and \textit{credulous} reasoning, in which a conclusion is considered valid if it is supported by at least one model. We sample 200 problems for each.
\textbf{BoardgameQA} \cite{kazemi23boardgameqa} evaluates defeasible reasoning over contradictory natural language information, where one must resolve conflicts by reasoning with rule preferences. We sample 200 problems.

\noindent
\textbf{Baseline LLMs.} We used Gemini 2.5 Pro, Gemini 2.5 Flash, o4-mini, and DeepSeek R1-0528 as Reasoning models and DeepSeek V3 as a non-Reasoning model (i.e., Instruct model).
These models were selected to represent a range of capabilities, from powerful, state-of-the-art models to more efficient ones, and to include both proprietary and open-weight options.
We report mean accuracy across 3 trials with standard deviations.

\subsection{RQ1: Overall Performance} \label{ssec:analysis-baseline}

\noindent
{\bf {\sf LLM+ASP} Substantially Outperforms Baseline LLMs, with Largest Gains on Hardest Problems.}\ \ 
Table~\ref{tab:main-result} presents our main results comparing baseline LLMs and {\sf LLM+ASP} with compact reference.

{\sf LLM+ASP} achieves 78.0\% average accuracy, a 52.1\% relative improvement over baselines (51.3\%). This is remarkably consistent: even strong o4-mini (63.4\%) reaches 82.3\% with ASP. Weaker models see dramatic gains—Gemini 2.5 Flash jumps from 43.3\% to 86.9\%.

Benefits are most pronounced on hardest problems. On \textbf{ZL-XXL}, baseline averages 24.2\%, but {\sf LLM+ASP} achieves 71.3\%—a 195\% relative improvement. On SudokuBench, baselines achieve 8.4\% versus 59.8\% for {\sf LLM+ASP}.

\smallskip\noindent 
{\bf Model Capability Still Matters, But Less.}\ \ 
While {\sf LLM+ASP} reduces performance gaps, stronger base models achieve better results. Gemini 2.5 Pro+ASP leads at 93.9\%, followed by Gemini 2.5 Flash+ASP (86.9\%), o4-mini+ASP (82.3\%), and DeepSeek-R1+ASP (81.9\%). 
DeepSeek V3, a non-reasoning model, only reaches 45.0\%, suggesting iterative self-correction works better with Reasoning models. 

Interestingly, adding ASP does not simply preserve the baseline ordering of models. For example, o4-mini outperforms Gemini 2.5 Flash in the baseline setting (63.4\% vs.\ 43.3\%), but Gemini 2.5 Flash+ASP surpasses o4-mini+ASP once the symbolic pipeline is introduced (86.9\% vs.\ 82.3\%). 
This suggests that the two settings reward different capabilities: the baseline favors direct reasoning, whereas {\sf LLM+ASP} places greater weight on symbolic translation, program repair, and effective use of solver feedback. 
A plausible explanation is that Gemini 2.5 Flash benefits more from the ASP pipeline because it is relatively stronger at code generation, while o4-mini’s advantage lies in direct reasoning.

\smallskip\noindent 
{\bf Benchmark-Specific Insights.}\ \ 

\noindent
{\bf (i) Constraint satisfaction:} \textbf{ZL-XL}, \textbf{ZL-XXL}, and \textbf{SudokuBench} show clearest advantages. 
On \textbf{ZL-XXL}, Gemini 2.5 Pro+ASP achieves 97.7\% including perfect performance on 12$\times$12 puzzles, while baseline drops to 0\%. Top three {\sf LLM+ASP} systems achieve 67--74.7\% on \textbf{SudokuBench} versus 8.3--15.0\% baselines.

\noindent
{\bf (ii) Planning:} \textbf{Mystery Blocksworld} shows moderate overall improvement (72.6\% vs. 64.2\%). However, Gemini 2.5 Flash improves from 66.3\% to 96.7\%, and Gemini 2.5 Pro+ASP achieves 98.3\%.

\noindent
{\bf (iii) Nonmonotonic reasoning:} \textbf{MultiLogicNMR} demonstrates ASP's strength: 90.5\% skeptical and 81.0\% credulous versus baseline's 59.0\% and 52.9\%. Gemini 2.5 Pro+ASP achieves perfect 100\% skeptical accuracy. 
BoardgameQA shows no overall gains (82.4\% vs. 84.7\% baseline), with small gains (92.0\% vs. 88.0\%) among Reasoning models only. We attribute this primarily to task structure: \textbf{BoardgameQA} involves single-model reasoning with explicit priority orderings stated directly in the input, allowing LLMs to learn an algorithmic conflict-resolution procedure through pattern matching.\footnote{There may also be a data contamination issue, as \textbf{BoardgameQA} was published in 2023, before the training data cutoffs of these LLMs.}
In contrast, \textbf{MultiLogicNMR} requires genuine nonmonotonic inference---distinguishing whether conclusions hold in \emph{all} stable models (skeptical) versus \emph{at least one} (credulous)---a capability that requires enumerating multiple possible world interpretations, which is challenging for autoregressive token prediction but naturally handled by ASP's stable model semantics.



\begin{figure*}[t!]
    \centering
   \includegraphics[width=1.0\linewidth,height=4.8cm]{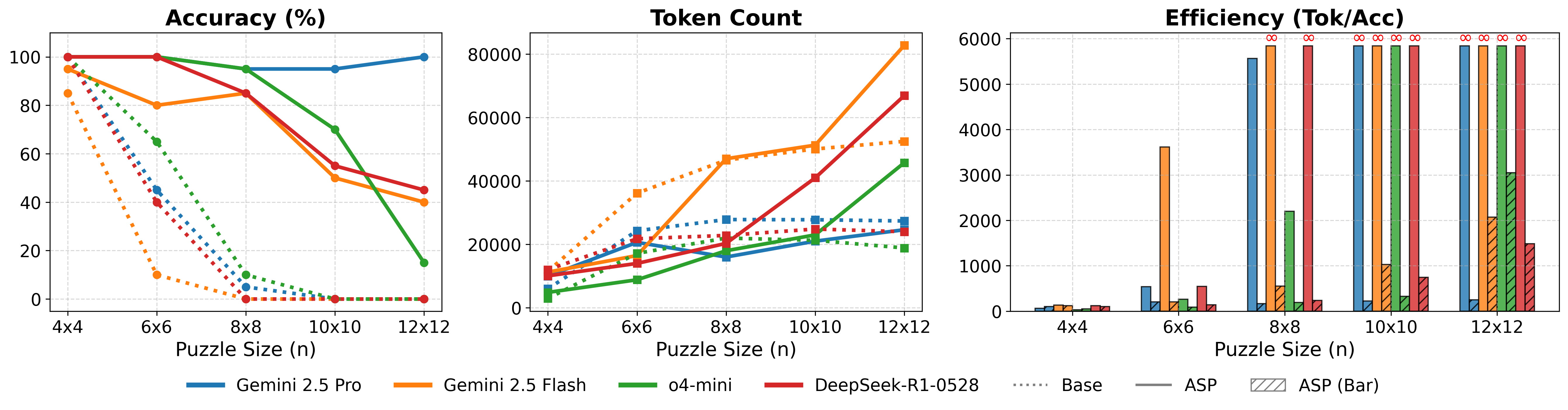}
    \caption{Reasoning Models' Performance vs. difficulty (left), output token usage vs. difficulty (middle), and the ratio of output token usage and accuracy vs. difficulty (right), on \textbf{ZL-XXL}.}
    \label{fig:zebra-combined}
\end{figure*}

\smallskip\noindent 
{\bf {\sf LLM+ASP} Maintains Effort on Hard Problems While Baseline LLMs Give Up.}\ \ 
Figure~\ref{fig:zebra-combined} reveals a striking pattern. On \textbf{ZL-XXL}, for baseline models, the output token usage peaks at 8$\times$8 problems then \textit{declines} on harder puzzles, coinciding with accuracy collapse---the models essentially give up. This aligns with findings that LLM performance breaks down past a critical complexity threshold \cite{shojaee25illusion, valmeekam25systematic}.

On the other hand, 
{\sf LLM+ASP} output token usage increases monotonically with difficulty, reflecting sustained engagement. Crucially, this is \textbf{efficient}: baseline LLMs consume tokens to directly traverse search space, while {\sf LLM+ASP} generates compact ASP programs that guide efficient symbolic search. The efficiency advantage is particularly evident at the highest difficulty: on 12$\times$12 problems, Gemini 2.5 Pro+ASP achieves 100\% accuracy using approximately 25,000 tokens, while baseline Gemini 2.5 Pro uses similar token counts to achieve 0\% accuracy. The rightmost panel shows that {\sf LLM+ASP} consistently maintains better token efficiency (lower tokens per correct solution) across all difficulty levels, with the gap widening dramatically as problem complexity increases.

\subsection{RQ2: Impact of In-Context Knowledge} 
\label{sec:rq2}
\begin{table}[t]
\centering
\fontsize{7}{8}\selectfont 
\caption{Impact of External Knowledge} \vspace{-1em}
\label{tab:llm-asp-reference-comparison}
\setlength{\tabcolsep}{3.5pt}
\begin{tabular}{l|c|c|c|c|c}
\toprule
\multirow{2}{*}{\textbf{Model}} & \multirow{2}{*}{\textbf{ZL-XL}} & \multirow{2}{*}{\textbf{ZL-XXL}} & \multicolumn{1}{c|}{\textbf{Sudoku-}} & \multirow{1}{*}{\textbf{Mystery-}} & \multirow{2}{*}{\textbf{Avg.}} \\
&  &  & \textbf{Bench} & \textbf{BW} &   \\
\midrule
\multicolumn{6}{c}{{{\sf LLM+ASP} with No Reference}} \\
\midrule

Gem. 2.5 Pro+ASP    & 97.0 & 85.0  & \textbf{76.0} & 69.0 & 81.8  \\
Gem. 2.5 Flash+ASP  & 38.0 & 25.0  & 66.0 & 66.0 & 48.8    \\
o4-mini+ASP           & 97.0 & 67.0  & 72.0 & 41.0 & 69.3  \\
DS-R1-0528+ASP  & 95.0 & 76.0  & 68.0 & 64.0 & 75.8 \\
DS-V3+ASP       & 49.0 & 28.0  & 22.0 & 26.0 & 31.3  \\
\midrule
Average & 75.2 & 56.2  & 60.8  &  53.2 & 61.4  \\
\midrule

\multicolumn{6}{c}{{{\sf LLM+ASP} with Standard Reference}} \\
\midrule

Gem. 2.5 Pro+ASP    & 95.0 & 94.0 & 75.0 & \textbf{98.0} &  90.5  \\
Gem. 2.5 Flash+ASP  & 89.0 & 54.0 & 71.0 & 96.0 &  77.5  \\
o4-mini+ASP           & \textbf{98.0} & 68.0 & 65.0 & 62.0 &  73.3  \\
DS-R1-0528+ASP  & \textbf{98.0} & 68.0 & 68.0 & 73.0 &  76.8  \\
DS-V3+ASP       & 63.0 & 24.0 & 26.0 & 15.0 &  32.0  \\
\midrule
Average & 88.6  &  61.6 & 61.0  & 68.8 & 70.0  \\
\midrule

\multicolumn{6}{c}{{{\sf LLM+ASP} with Compact Reference}} \\
\midrule

Gem. 2.5 Pro+ASP    & 96.7 & \textbf{97.7}  & 74.7  & \textbf{98.3}  & \textbf{91.9} \\
Gem. 2.5 Flash+ASP  & 84.3 & 70.3 & 65.7 & 96.7  &  79.3 \\
o4-mini+ASP           & 95.0 & 76.7  & 67.0 & 72.7  & 77.9 \\
DS-R1-0528+ASP  & \textbf{97.3} &  75.0 & 70.0  & 81.7  & 81.0 \\
DS-V3+ASP       & 69.0 &  37.0 & 21.7 & 13.7  &  35.4 \\
\midrule
Average & 88.5  & 71.3   & 59.8  & 72.6 & 73.1  \\
\bottomrule 
\end{tabular}
\end{table}

\noindent
{\bf Evidence Shows an ASP Blind Spot in LLMs; In-Prompt Reference Material Mitigates This.}\ \ 
To assess the impact of external knowledge on performance, we experimented with the {\sf LLM+ASP} framework with and without references.

As shown in Table~\ref{tab:llm-asp-reference-comparison}, even without any reference, {\sf LLM+ASP} already clearly outperforms the baseline LLMs (Table~\ref{tab:main-result}).
However, we find that LLMs exhibit persistent errors in generating ASP programs, such as operator misuse (e.g., \texttt{abs}), faulty pooling constructs, and flawed modeling of actions and effects. These errors motivate the use of in-prompt ASP references. 

Overall, incorporating an ASP reference substantially improves performance over using no reference. Averaged across all models and \textbf{ZL-XL}, \textbf{ZL-XXL}, \textbf{SudokuBench}, and \textbf{Mystery Blocksworld}, accuracy increases from 61.4\% without a reference to 70.0\% when a standard reference is provided. Importantly, the compact reference consistently outperforms the standard reference, achieving 73.1\%, despite being only about one-eighth the length. 

The improvement is most pronounced for Gemini 2.5 Flash. With the compact reference, Gemini 2.5 Flash+ASP's overall accuracy jumps from 48.8\% to 79.3\%, approaching the performance levels of DeepSeek-R1+ASP (81.0\%) and surpassing o4-mini+ASP (77.9\%). 

The compact reference is more effective, not because it introduces new information, but because it selectively emphasizes the frequent ASP constructs. This effect is reflected quantitatively: on \textbf{ZL-XXL}, reasoning models collectively exhibit 84 syntax errors without any reference, 41 errors with the standard reference, and only 16 errors when using the compact reference (Appendix~\ref{appendix:error-analysis-zl-xxl}). By distilling core conventions and modeling patterns into succinct form, the compact reference more reliably guides the model toward correct operator usage, pooling, and action modeling. A similar trend appears for semantic omissions: for reasoning models collectively, missing constraints increase from 46 without a reference to 66 with the longer standard reference, but drop to 32 with the compact reference. This identifies a ``context rot'' phenomenon where verbose documentation hinders the LLM's ability to attend to specific constraints, aligned with prior work showing that excessive context can hinder LLM retrieval and reasoning \cite{kuratov24babilong, chatziveroglou25exploring, hengle25can, hong25context}.

\subsection{RQ3: Impact of Iterative Correction}  \label{ssec:rq3}

\newcommand{\subhead}{\makecell{w/o\\rev.}}
\newcommand{\subheadrev}{\makecell{with\\rev.}}
\newcommand{\subheadavg}{\makecell{avg\\\#}}

\begin{table*}[ht]
    \centering
    \caption{{\sf LLM+ASP} Performance with compact reference.} \vspace{-1em}
    \footnotesize
    \setlength{\tabcolsep}{4pt}
    \label{tab:revision-comparison}
    \begin{tabular}{lcccccccccccc}
        \toprule
        \multirow{2}{*}{\textbf{LLM Model}} & \multicolumn{3}{c}{\textbf{ZL-XL (\%)}} & \multicolumn{3}{c}{\textbf{ZL-XXL (\%)}} & \multicolumn{3}{c}{\textbf{SudokuBench (\%)}} & \multicolumn{3}{c}{\textbf{MysteryBW (\%)}} \\
        \cmidrule(lr){2-4} \cmidrule(lr){5-7} \cmidrule(lr){8-10} \cmidrule(lr){11-13}
        & \subhead & \subheadrev & \subheadavg & \subhead & \subheadrev & \subheadavg & \subhead & \subheadrev & \subheadavg & \subhead & \subheadrev & \subheadavg \\
        \midrule
        Gemini 2.5 Pro & 86.0 & 96.7 & 0.19 & 78.0 & 97.7 & 0.53 & 61.0 & 74.7 & 1.50 & 86.0 & 98.3 & 0.23 \\
        Gemini 2.5 Flash & 49.0 & 84.3 & 0.67 & 38.0 & 70.3 & 1.28 & 38.0 & 65.7 & 2.30 & 70.0 & 96.7 & 0.77 \\
        o4-mini & 39.0 & 95.0 & 0.90 & 18.0 & 76.7 & 2.76 & 17.0 & 67.0 & 3.48 & 13.0 & 72.7 & 3.21 \\
        DS-R1-0528 & 56.0 & 97.3 & 0.65 & 49.0 & 75.0 & 1.71 & 33.0 & 70.0 & 3.34 & 50.0 & 81.7 & 1.25 \\
        DS-V3 & 19.0 & 69.0 & 3.04 & 7.0 & 37.0 & 5.12 & 9.0 & 21.7 & 6.93 & 1.0 & 13.7 & 7.76 \\
        \textbf{Average} & \textbf{49.8} & \textbf{88.5} & \textbf{1.09} & \textbf{38.0} & \textbf{71.3} & \textbf{2.28} & \textbf{31.6} & \textbf{59.8} & \textbf{3.51} & \textbf{44.0} & \textbf{72.6} & \textbf{2.64} \\
        
        \midrule \midrule
        
        \multirow{2}{*}{\textbf{LLM Model}} & \multicolumn{3}{c}{\textbf{Skeptical}} & \multicolumn{3}{c}{\textbf{Credulous}} & \multicolumn{3}{c}{\textbf{BoardgameQA}} & \multicolumn{3}{c}{\textbf{Overall Average}} \\
        \cmidrule(lr){2-4} \cmidrule(lr){5-7} \cmidrule(lr){8-10} \cmidrule(lr){11-13}
        & \subhead & \subheadrev & \subheadavg & \subhead & \subheadrev & \subheadavg & \subhead & \subheadrev & \subheadavg & \subhead & \subheadrev & \subheadavg \\
        \midrule
        Gemini 2.5 Pro & 99.5 & 100 & 0.03 & 97.5 & 93.8 & 0.19 & 78.0 & 96.3 & 0.50 & 83.7 & 93.9 & 0.45 \\
        Gemini 2.5 Flash & 98.0 & 98.3 & 0.02 & 97.0 & 98.3 & 0.08 & 65.5 & 94.7 & 0.91 & 65.1 & 86.9 & 0.86 \\
        o4-mini & 58.0 & 95.8 & 0.96 & 59.5 & 84.0 & 0.97 & 51.5 & 85.2 & 1.41 & 36.6 & 82.3 & 1.96 \\
        DS-R1-0528 & 76.0 & 83.8 & 2.46 & 76.5 & 73.5 & 1.99 & 53.5 & 91.7 & 3.95 & 56.3 & 81.9 & 2.19 \\
        DS-V3 & 74.3 & 74.3 & 3.33 & 71.0 & 55.3 & 4.59 & 39.5 & 44.0 & 7.27 & 31.5 & 45.0 & 5.43 \\
        \textbf{Average} & \textbf{80.7} & \textbf{90.5} & \textbf{1.36} & \textbf{80.3} & \textbf{81.0} & \textbf{1.56} & \textbf{57.6} & \textbf{82.4} & \textbf{2.80} & \textbf{54.6} & \textbf{78.0} & \textbf{2.18} \\
        \bottomrule
    \end{tabular}
\end{table*}


\noindent
{\bf Iterative Self-Correction is the Primary Driver of Performance.}\ \ 
An analysis of Table~\ref{tab:revision-comparison} reveals that the iterative revision process is the single most crucial factor for the success of the {\sf LLM+ASP} framework, with performance jumping from an average of {54.6\%} without revisions to {78.0\%} when models are allowed to self-correct. 
Note that even without any revisions, {\sf LLM+ASP} already outperforms standalone baselines, confirming that LLMs can generate reasonable ASP programs from natural language descriptions---though the self-correction loop remains the primary driver of the full performance gains. The average number of revisions was 2.18, and the average runtime was only 2.3 seconds.

Furthermore, the results highlight a clear hierarchy in model efficiency: a top-tier model like Gemini 2.5 Pro not only achieves the highest accuracy at 93.9\% but does so with remarkable efficiency, requiring less than one revision on average (0.45). In stark contrast, a non-Reasoning model like DeepSeek V3 needs over five revisions (5.43) on average. This underscores that while the self-correction loop is universally beneficial, stronger Reasoning models exploit it much more effectively.

To further characterize verifier reliability, Table~\ref{tab:incorrect-pass-rate} reports the false-accept rate: among cases where the pipeline issued a PASS decision, the fraction for which the approved output was actually incorrect. Denominators differ across models because they reflect only approved outputs, excluding cases that never reached approval within the revision budget (due to persistent unsatisfiability, compilation errors, timeouts, or LLM rejection).
The rates vary substantially: Gemini 2.5 Pro (5.06\%), Gemini 2.5 Flash (8.76\%), o4-mini (12.42\%), DeepSeek R1-0528 (13.86\%), and DeepSeek V3 (37.33\%). Stronger models serve as more reliable verifiers. DeepSeek V3's high rate is especially notable given its smaller denominator: it approves fewer outputs overall, yet still errs on over a third of them—indicating weakness in both generation and verification.

\begin{table}[h!]
\centering
\small
\caption{
False-accept rate across models (with compact reference, aggregated over all 1000 problems). Denominator counts only instances where the pipeline issued a final PASS decision; numerator counts approved outputs that are actually incorrect.}
\label{tab:incorrect-pass-rate}
\setlength{\tabcolsep}{4pt}
\begin{tabular}{lc}
\toprule
\textbf{Model} & \textbf{Incorrect PASS Rate} \\
\midrule
{Gemini 2.5 Pro}   & {5.06\%} (50/989) \\
{Gemini 2.5 Flash} & {8.76\%} (86/982) \\
{o4-mini}          & {12.42\%} (119/958) \\
{DS-R1-0528} & {13.86\%} (130/938) \\
{DS-V3}      & {37.33\%} (249/667) \\

\bottomrule
\end{tabular}
\end{table}


\begin{table}[h!]
\centering
\fontsize{8}{9}\selectfont 
\caption{Num. of Errors in the final ASP code produced by {\sf LLM+ASP} (with reference) for all reasoning models.} \vspace{-1em}
\label{tab:errors-with-reference}
\setlength{\tabcolsep}{3pt}
\begin{tabular}{llcccc}
\toprule
\textbf{Mode} & \textbf{Benchmark} & \textbf{Syntax} & \textbf{Timeout} & \textbf{Sat} & \textbf{Unsat} \\
\midrule
{No} & \textbf{ZL-XL} & 79 & 0 & 27 & 23 \\
{Revisions} & \textbf{ZL-XXL} & 123 & 15 & 54 & 42\\
{} & \textbf{SudokuBench} & 158 & 2 & 36 & 51 \\
\cmidrule{2-6}
& {\bf Combined} & \textbf{360} & \textbf{17} & \textbf{117} & \textbf{116}\\
&  &  (59.0\%) & (2.8\%) & (19.2\%) & (19.0\%) \\
\midrule
{With} & \textbf{ZL-XL} & 0 & 0 & 4 & 16 \\
{Revisions} & \textbf{ZL-XXL} & 7 & 4 & 58 & 47\\
{} & \textbf{SudokuBench} & 11 & 5 & 65 & 41\\
\cmidrule{2-6}
& \textbf{Combined} & \textbf{18} & \textbf{9} & \textbf{127} & \textbf{104}\\
& & (7.0\%) & (3.5\%) & (49.2\%)  & (40.3\%) \\
\bottomrule
\end{tabular}
\end{table}

\smallskip
\noindent
{\bf An Analysis of Failure Modes of {\sf LLM+ASP}.}\ \ 
We classify the error cases of the final ASP programs from \textbf{ZL-XL}, \textbf{ZL-XXL}, and \textbf{SudokuBench} in Table~\ref{tab:errors-with-reference} into four types: {\bf Syntax} (the program fails to compile), {\bf Timeout} (the program does not terminate within the 80-second limit), {\bf Sat} (the program produces an answer set but it is incorrect) and {\bf Unsat} (the program is unsatisfiable).

When no revisions are allowed, the majority of errors in the generated ASP programs are syntax-related, with 360 such cases. However, when revisions are allowed, syntax errors are drastically reduced to just 18 cases. 
Once the syntax issues are resolved, the majority of remaining errors are semantic in nature, involving programs that are either unsatisfiable or produce incorrect outputs. Interestingly, the absolute number of satisfiable but incorrect programs increases slightly with revisions, likely because once syntax issues are largely addressed, remaining semantic flaws emerge that were previously masked.

The two types of semantic errors differ in their debuggability: unsatisfiable programs provide minimal solver feedback (no answer sets), making it difficult for the LLM to identify the underlying logic flaws. This issue could be alleviated by incorporating unsat-core extraction~\citep{alviano18cautious} or off-the-shelf ASP debugging tools (e.g., {\sc spock}~\citep{gebser07spock}, {\sc Ouroboros}~\citep{oetsch10catching}), which we leave as future work. In contrast, satisfiable but incorrect programs yield concrete answer sets that can be checked against the problem description, offering the LLM clearer information for correction.

\subsection{RQ4: Experiments with {\sf LLM+SMT}} \label{ssec:rq4}

To investigate how the choice of symbolic formalism affects performance, we modify our framework to use SMT in place of ASP using the same self-correction pipeline. We call this {\sf LLM+SMT} and compare with {\sf LLM+ASP} on the same benchmarks.
Prior work~\citep{hao25large} achieved strong SMT results using extensive task-specific engineering (10+ handcrafted prompts per domain, few-shot examples, domain schemas). To isolate the impact of the symbolic formalism itself rather than engineering effort, we compare {\sf LLM+ASP} and {\sf LLM+SMT} in equivalent zero-shot settings without references.

For fair comparison, we provide no reference materials to either system in these experiments. Table~\ref{tab:llm-smt-preliminary} presents the results, revealing two key findings:

\begin{table*}[t]
\centering
\fontsize{8}{9}\selectfont 
\caption{{\sf LLM+ASP} vs. {\sf LLM+SMT} with Revisions (No Reference)}
\label{tab:llm-smt-preliminary}
\setlength{\tabcolsep}{3.5pt}
\begin{tabular}{l|c|c|c|c|c|c|c|c}
\toprule
\textbf{Model} & \textbf{ZL-XL} & \textbf{ZL-XXL} & \textbf{Sudoku} & \textbf{MysteryBW} & \textbf{Skeptical} & \textbf{Credulous} & \textbf{BoardQA}  & \textbf{Average} \\
 & (\%) & (\%) & (\%) & (\%) & (\%) & (\%) & (\%) & (\%) \\
 \midrule
\multicolumn{9}{c}{\textbf{LLM + ASP}} \\
\midrule
Gemini 2.5 Pro+ASP  &97.0 & 85.0 & 76.0 & 69.0 & 100.0 & 96.5 & 96.0 & 88.5 \\
Gemini 2.5 Flash+ASP &38.0 & 25.0 & 66.0 & 66.0 & 99.5 & 98.5 & 92.5 & 69.4 \\
o4-mini+ASP     &97.0 & 67.0 & 72.0 & 41.0 & 96.5 & 82.0 & 91.0 & 78.1 \\
DS-R1-0528+ASP &95.0 & 76.0 & 68.0 & 64.0 & 91.5 & 87.5 & 89.5 & 81.6 \\
DS-V3+ASP     &49.0 & 28.0 & 22.0 & 26.0 & 77.0 & 57.0 & 41.5 & 42.9 \\
\midrule
Average &75.2 & 56.2 & 60.8 & 53.2 & 92.9 & 84.3 & 82.1 & 72.1 \\
\midrule
\multicolumn{9}{c}{\textbf{LLM + SMT}} \\
\midrule
Gemini 2.5 Pro+SMT    & 98.0 & 100.0 & 74.0 & 98.0 & 32.0 & 37.0 & 52.0 & 70.1 \\
Gemini 2.5 Flash+SMT    & 97.0 & 88.0 & 68.0 & 65.0 & 36.0 & 36.0 & 55.0 & 63.6 \\
o4-mini+SMT    & 92.0 & 84.0 & 56.0 & 55.0 & 36.5 & 34.5 & 80.0 & 62.6 \\
DS-R1-0528+SMT    & 94.0 & 80.0 & 59.0 & 76.0 & 36.5 & 33.0 & 49.5 & 61.1 \\
DS-V3+SMT    & 39.0 & 9.0 & 43.0 & 9.0 & 38.0 & 34.5 & 64.0 & 33.8 \\
\midrule
Average & 84.0 & 72.2 & 60.0 & 60.6 & 35.8 & 35.0 & 60.1 & 58.2 \\
\bottomrule
\end{tabular}
\end{table*}

\smallskip
\noindent
\textbf{(1) {\sf LLM+SMT} Does Well on Constraint Satisfaction and Planning Benchmarks.}\ \
On constraint satisfaction benchmarks, {\sf LLM+SMT} outperforms {\sf LLM+ASP}: \textbf{ZL-XL} (84.0\% vs. 75.2\%), \textbf{ZL-XXL} (72.2\% vs. 56.2\%), and \textbf{Mystery Blocksworld} (60.6\% vs. 53.2\%). \textbf{SudokuBench} shows comparable performance (60.0\% vs. 60.8\%). We attribute this advantage primarily to training data exposure: LLMs likely have encountered substantially more SMT code than ASP programs during pretraining, as Z3's Python bindings are widely used. However, {\sf LLM+ASP} without reference is prone to generating more erroneous ASP programs. Interestingly, most of these issues are alleviated when using the compact reference described in Section~\ref{sec:rq2}, making {\sf LLM+ASP} with the reference perform better than {\sf LLM+SMT} (with no reference). 

\smallskip
\noindent
{\bf (2) {\sf LLM+SMT} is Not Suitable for Nonmonotonic Reasoning Benchmarks.}\ \ 
The performance pattern is reversed on tasks requiring nonmonotonic reasoning. On \textbf{MultiLogicNMR}, {\sf LLM+ASP} achieves 92.9\% (skeptical) and 84.3\% (credulous), while {\sf LLM+SMT} reaches only 35.8\% and 35.0\%, barely above random chance (33.3\%). Similarly, on \textbf{BoardgameQA}, {\sf LLM+ASP} achieves 82.1\% compared to 60.1\% for {\sf LLM+SMT}. Indeed, {\sf LLM+SMT} performed even worse than the standalone LLMs: 59.0\% skeptical and 52.9\% credulous on \textbf{MultiLogicNMR}, and 84.7\% on \textbf{BoardgameQA} (c.f. Table~\ref{tab:main-result}).

Analysis of {\sf LLM+SMT} outputs reveals that the model frequently responds with ``unknown'' on the \textbf{MultiLogicNMR} skeptical task ({94.5\%} of problems) and \textbf{BoardgameQA} (72.5\% of problems), far exceeding the correct answer rate of 33\%. This failure pattern stems from a fundamental mismatch: SMT solvers operate under monotonic logic, where adding information can only derive new conclusions, never retract them. Default negation—the ability to assume something is false unless proven true—cannot be directly expressed in SMT.\footnote{For example, in ``A if B unless C,'' unless one can prove C, C is assumed to be false by default.} 

ASP, by contrast, was designed explicitly for nonmonotonic reasoning via stable model semantics. Default rules, exceptions, and alternative models are first-class constructs in ASP, allowing LLMs to express these patterns naturally with minimal guidance.

\smallskip
\noindent
{\bf Addressing Limitations of Prior Approaches.}\ \ 
A key barrier to practical neuro-symbolic reasoning has been the engineering effort required to couple LLMs with formal solvers. Our {\sf LLM+SMT} results also compare favorably to prior work coupling LLMs with SMT solvers. \citet{hao25large} achieved strong results on constraint satisfaction tasks, but their approach required extensive task-specific engineering efforts: up to 10 handcrafted prompts with few-shot examples per domain, and domain-specific schemas.

When we apply their methodology in a zero-shot setting (see Appendix~\ref{app:hao}), performance degrades substantially. Manual inspection reveals persistent encoding errors including incorrect variable scoping and malformed constraints. Our {\sf LLM+SMT} framework, utilizing the same self-correction mechanism that enables effective ASP generation, achieves better results without any task-specific prompt engineering. 

\section{Conclusion}
This paper introduced {\sf LLM+ASP}, a neuro-symbolic framework that integrates the natural language processing capabilities of LLMs with the formal reasoning of ASP. Our approach demonstrates that recent Reasoning models can effectively translate complex problems into ASP programs, which are then solved by an ASP solver. The framework's core strength lies in an iterative self-correction loop where the structured feedback from the ASP solver enables the Reasoning model to refine its generated programs.

We find this self-correction capability, further enhanced by providing the LLM with a compact ASP reference guide, is the primary driver of performance gains, yielding substantial improvements over baseline LLMs (52\% relative improvement on average). For nonmonotonic reasoning tasks involving defaults and exceptions, ASP's theoretical foundation provides inherent advantages that are difficult to replicate in monotonic formalisms, as evidenced by the substantial performance gaps on such benchmarks.

Though LLMs can effectively produce ASP programs for complex problems, the gains from using reference materials suggest that better LLM training which includes more ASP-specific material could yield substantial performance improvements. We envision a future where LLMs act as flexible interfaces to a variety of formal reasoners, efficiently delegating complex logical tasks to broaden their scope of application.

\section*{Limitations}

\paragraph{Model Capability Dependency.} The framework's performance depends strongly on the underlying LLM's reasoning capabilities. While stronger models like Gemini 2.5 Pro achieve high accuracy with fewer than one revision on average, non-reasoning models like DeepSeek V3 require over five revisions while still achieving much lower success rates, suggesting the self-correction mechanism cannot fully compensate for weaker base models.


\paragraph{Debugging Unsatisfiable Programs.} When a generated program produces no answer sets, the solver provides minimal feedback, making it difficult for the LLM to identify logical contradictions. Future work could incorporate unsat-core extraction or dedicated ASP debugging tools (e.g., \textsc{Spock}, \textsc{Ouroboros}) to provide more actionable feedback.

\paragraph{Computational Costs.} Our reported experimental costs reflect approximately \$3,800 in direct LLM API charges for the core experiments, though the total expenditure, including preliminary testing, failed runs, and development, is estimated at \$4,200. 

\paragraph{Limited Scope.} Our evaluation focuses on constraint satisfaction, planning, and nonmonotonic reasoning tasks. Applicability to broader NLP tasks and more ambiguous or open-ended problems remains an open question.

\section*{Acknowledgements}
We are grateful to the anonymous referees for their useful comments.

\bibliography{bib,bib2,bib-gpt-action,bib-llm-asp}

~
\newpage 
~
\newpage 

\appendix

\setcounter{secnumdepth}{2}


{\bf \Large Appendix}
\medskip

The appendix is organized as follows:
Section~\ref{appendix:experiments-details} has the details of the experiments settings.
Section~\ref{appendix:prompts} contains the prompts used for the generator and formatter components, as illustrated in Figure~\ref{fig:main-figure}.
Section~\ref{appendix:example} provides a representative example of the LLM+ASP pipeline to help the reader understand how each component and step works in practice.   
Section \ref{appendix:error-analysis-zl-xxl} presents an expanded analysis of the effect of the reference, showing how different reference materials (none, standard, compact) affect common error types such as operator misuse, pooling issues, and missing constraints.
Section~\ref{appendix:iterative-corrections-no-reference} provides expanded analysis of iterative corrections without references.
Section~\ref{appendix:expanded-tables} contains expanded tables that show the detailed error distribution of LLM+ASP for each model and benchmark, categorized by error type (syntax, timeout, satisfiable-but-incorrect, unsatisfiable), with breakdowns both with and without references, and with and without revisions.
Section~\ref{app:hao} presents preliminary comparison with our {\sf LLM+ASP} with the method by \cite{hao25large}.
Section~\ref{appendix:compact-reference}
 provides the complete compact reference material used in the experiments.

\section{Experiments Settings Details} \label{appendix:experiments-details}

All experiments were performed on a machine with an Intel Core i9-9900K CPU, an NVIDIA GeForce RTX 2080 Ti GPU, and 32 GB of RAM. 
LLM inference was done via API; local hardware was used only for running {\sc clingo}.
Experiments were conducted in November to December 2025. 
We allowed a maximum of 10 revisions, which was more than sufficient. The average number of revisions was 2.18, and we use a timeout limit of 80 seconds. 
The temperature parameter used for all LLMs was the default, 1.0.

The datasets used in this work are publicly available and distributed under the following licenses: ZebraLogicBench (\textbf{ZL-XL}) under the Apache License 2.0, while the newly introduced \textbf{ZL-XXL} is released under the CC BY 4.0 license,
\textbf{SudokuBench} and \textbf{Mystery Blocksworld} under the MIT License, \textbf{BoardgameQA} under the CC-BY 4.0 license, while \textbf{MultiLogicNMR} does not specify a license in its original release.

\BOCC
The appendix is organized as follows. Section~\ref{appendix:prompts} contains the prompts used for the generator and formatter, as illustrated in Figure~\ref{fig:main-figure}. Section \ref{appendix:example} contains a representative example of {\sf LLM+ASP}, to help the reader understand how each component and step works. 
Section~\ref{appendix:expanded-tables} contains expanded tables which show the error distribution of {\sf LLM+ASP} for each model, as well as including the distribution for without a reference. All experiments were performed on a machine with an Intel Core i9-9900K CPU, an NVIDIA GeForce RTX 2080 Ti GPU, and 32 GB of RAM.
\EOCC{}

\section{Prompts} \label{appendix:prompts}


\subsection{Generator Prompt} \label{sec:app:generator-prompt}
The prompt for the generator block as shown in Figure~\ref{fig:main-figure} is as follows.
\begin{lstlisting}[escapechar=^]
Consider the following resource material describing Answer Set Programming and the language of Clingo to do the following task.
<resource-material>
^{\cblu \textlangle resource omitted\textrangle}^ 
</resource-material>


<instructions>
You are tasked with constructing an ASP program to solve the following problem:
<problem-description>
^{\cblu \textlangle input problem description\textrangle}^ 
IMPORTANT: Atoms with large arity may take very long for larger problems, so do not introduce them unnecessarily.
</problem-description>
The ASP program should be made of ASP modules, which each make up a necessary part of the program to represent the problem.
In the program, include the #show command for atoms which will help for verifying the answer and/or debugging. For example, don't just show the final answer, but show atoms which can be used to check that the parts determining the final answer are also correct.

<current-clingo-modules>
^{\cblu \textlangle previously generated ASP program omitted, see above \textrangle}^ 
</current-clingo-modules>
<candidate-answer-set>
^{\cblu \textlangle clingo solver output \textrangle}^ 
</candidate-answer-set>


You are tasked with doing an operation from the following.
(UPDATE) - This should be done either to:
	(a) write an initial ASP program (still considered an update).
	(b) update the program based on the Clingo output which may help to debug the current output.
If updated, make sure to write the complete program.
(PASS) - This should be done when the Clingo output is correct. There should be no ambiguity and consensus on the proposed solution being correct. ONLY use when absolutely sure the output is correct. If the proposed solution is correct, you may ignore stderr and still pass.

For UPDATE, format your output exactly like the following, writing the complete program:

```
% MODULES START
% module <name of first module>
<ASP code for this module>
% module <name of first module> END

% module <name of second module>
<ASP code for this module>
% module <name of second module> END
...
% MODULES END
```
For PASS, format your output exactly like the following (only write the action, since nothing else is to be done):

```
OPERATION: PASS
```

Do not write anything outside of the three backticks.
</instructions>
\end{lstlisting}

\subsection{Human-Readable Formatting Prompt}

The prompt for the formatter block as shown in Figure~\ref{fig:main-figure} is as follows.

\begin{lstlisting}[escapechar=^]
You are tasked with writing the answer set produced by running Clingo into a human-readable form.

Here is the problem description:
^{\cblu \textlangle problem description \textrangle}^ 

Clingo output:
^{\cblu \textlangle clingo solver output \textrangle}^ 
<INSTRUCTION>
\end{lstlisting}

\section{{\sf LLM+ASP} Example}\label{appendix:example}

\textbf{Input Problem}
This problem is a $6 \times 6$ \textbf{ZL-XL} puzzle from \citet{richardson25zebra}.
\begin{tcolorbox}[breakable,fontupper=\small, title=Example Problem Input, label=input_ex]
There are 6 houses, numbered 1 to 6 from left to right, as seen from across the street. Each house is occupied by a different person. Each house has a unique attribute for each of the following characteristics:

house types: farmhouse, cabin, studio, apartment, loft, townhouse

nationalities: norwegian, japanese, italian, chinese, mexican, indian

cigarette brands: prince, camel, lucky strike, benson \& hedges, marlboro, l\&m

fictional genre: satire, sci-fi, dystopian, fairy tale, drama, war

clothing item: vest, belt, jeans, t-shirt, coat, watch

occupations: lawyer, actor, astronaut, magician, mechanic, architect

Clues:

There are two houses in between the person who smokes lucky strike and the person whose favorite fictional genre is drama.

The person who smokes prince is directly to left of the magician.

There are two houses in between the person who smokes lucky strike and the person who wears the vest.

The person in the cabin and the mechanic are the same person.

The lawyer is in house number 1.

The indian person and the person whose favorite fictional genre is war are the same person.

There are two houses in between the person who smokes benson \& hedges and the person whose favorite fictional genre is dystopian.

The person whose favorite fictional genre is satire is adjacent to the person whose favorite fictional genre is fairy tale.

The person whose favorite fictional genre is fairy tale is somewhere to the right of the architect.

The person who wears the vest is somewhere to the right of the person whose favorite fictional genre is sci-fi.

The mexican person is in house number 1.

The italian person and the person whose favorite fictional genre is dystopian are the same person.

The person in the farmhouse is directly to left of the person in the apartment.

The chinese person and the actor are the same person.

The norwegian person is adjacent to the astronaut.

The japanese person is directly to right of the person in the studio.

The person in the farmhouse is adjacent to the person whose favorite fictional genre is sci-fi.

There are two houses in between the person who smokes marlboro and the person who wears the t-shirt.

The person whose favorite fictional genre is satire is somewhere to the left of the person who smokes camel.

The person who smokes l\&m is not the person who wears the belt.

The astronaut is directly to left of the person in the apartment.

The person who wears the jeans is in house number 4.

The person in the townhouse is somewhere to the right of the person who wears the belt.

There is one house in between the person in the loft and the indian person.

The person who wears the jeans is adjacent to the person who wears the watch.

Find an assignment which satisfies all constraints in the problem.

\end{tcolorbox}

\subsection{1st Pass}\label{ssec:app:1st}

On the first pass through the generator, the problem and reference are included in the following prompt. Since this is the first generation, there is no previous program and the LLM will always generate the initial program. This is enforced since the only option available to the LLM is writing the initial program.

\subsubsection{Input} 

\begin{lstlisting}[escapechar=^]
Consider the following resource material describing Answer Set Programming and the language of Clingo to do the following task.
<resource-material>
^{\cblu \textlangle resource omitted\textrangle}^ 
</resource-material>


<instructions>
You are tasked with constructing an ASP program to solve the following problem:
<problem-description>
^{\cblu \textlangle input problem description omitted, see above\textrangle}^
IMPORTANT: Atoms with large arity may take very long for larger problems, so do not introduce them unnecessarily.
</problem-description>
The ASP program should be made of ASP modules, which each make up a necessary part of the program to represent the problem.
In the program, include the #show command for atoms which will help for verifying the answer and/or debugging. For example, don't just show the final answer, but show atoms which can be used to check that the parts determining the final answer are also correct.



You are tasked with doing an operation from the following.
(UPDATE) - This should be done either to:
	(a) write an initial ASP program (still considered an update).
	(b) update the program based on the Clingo output which may help to debug the current output.
If updated, make sure to write the complete program.

For UPDATE, format your output exactly like the following, writing the complete program:

```
% MODULES START
% module <name of first module>
<ASP code for this module>
% module <name of first module> END

% module <name of second module>
<ASP code for this module>
% module <name of second module> END
...
% MODULES END
```

Do not write anything outside of the three backticks.
</instructions>
\end{lstlisting}

\subsubsection{Output}
The generator produces the following program.
\begin{lstlisting}[escapechar=^]
```
% MODULES START
% module domain
% Houses
house(1..6).

% Categories and their unique attributes
house_type(farmhouse; cabin; studio; apartment; loft; townhouse).
nationality(norwegian; japanese; italian; chinese; mexican; indian).
cigarette(prince; camel; lucky_strike; benson_hedges; marlboro; l_m).
genre(satire; sci_fi; dystopian; fairy_tale; drama; war).
clothing(vest; belt; jeans; t_shirt; coat; watch).
occupation(lawyer; actor; astronaut; magician; mechanic; architect).
% module domain END

% module generate
% Generate a unique assignment for each attribute to a house.
% For each category, this creates a one-to-one mapping between the
% attributes of that category and the houses.

% house_is(House, Type)
{ house_is(H, T) : house(H) } = 1 :- house_type(T).

% person_is(House, Nationality)
{ person_is(H, N) : house(H) } = 1 :- nationality(N).

% smokes(House, Brand)
{ smokes(H, C) : house(H) } = 1 :- cigarette(C).

% likes(House, Genre)
{ likes(H, G) : house(H) } = 1 :- genre(G).

% wears(House, Item)
{ wears(H, I) : house(H) } = 1 :- clothing(I).

% works_as(House, Occupation)
{ works_as(H, O) : house(H) } = 1 :- occupation(O).
% module generate END

% module constraints
% This module translates the clues into integrity constraints.

% Clue 1: There are two houses in between the person who smokes lucky strike and the person whose favorite fictional genre is drama.
:- smokes(H1, lucky_strike), likes(H2, drama), |H1-H2| != 3.

% Clue 2: The person who smokes prince is directly to left of the magician.
:- smokes(H1, prince), works_as(H2, magician), H1 != H2-1.

% Clue 3: There are two houses in between the person who smokes lucky strike and the person who wears the vest.
:- smokes(H1, lucky_strike), wears(H2, vest), |H1-H2| != 3.

% Clue 4: The person in the cabin and the mechanic are the same person.
:- house_is(H, cabin), not works_as(H, mechanic).

% Clue 5: The lawyer is in house number 1.
:- not works_as(1, lawyer).

% Clue 6: The indian person and the person whose favorite fictional genre is war are the same person.
:- person_is(H, indian), not likes(H, war).

% Clue 7: There are two houses in between the person who smokes benson & hedges and the person whose favorite fictional genre is dystopian.
:- smokes(H1, benson_hedges), likes(H2, dystopian), |H1-H2| != 3.

% Clue 8: The person whose favorite fictional genre is satire is adjacent to the person whose favorite fictional genre is fairy tale.
:- likes(H1, satire), likes(H2, fairy_tale), |H1-H2| != 1.

% Clue 9: The person whose favorite fictional genre is fairy tale is somewhere to the right of the architect.
:- likes(H1, fairy_tale), works_as(H2, architect), H1 <= H2.

% Clue 10: The person who wears the vest is somewhere to the right of the person whose favorite fictional genre is sci-fi.
:- wears(H1, vest), likes(H2, sci_fi), H1 <= H2.

% Clue 11: The mexican person is in house number 1.
:- not person_is(1, mexican).

% Clue 12: The italian person and the person whose favorite fictional genre is dystopian are the same person.
:- person_is(H, italian), not likes(H, dystopian).

% Clue 13: The person in the farmhouse is directly to left of the person in the apartment.
:- house_is(H1, farmhouse), house_is(H2, apartment), H1 != H2-1.

% Clue 14: The chinese person and the actor are the same person.
:- person_is(H, chinese), not works_as(H, actor).

% Clue 15: The norwegian person is adjacent to the astronaut.
:- person_is(H1, norwegian), works_as(H2, astronaut), |H1-H2| != 1.

% Clue 16: The japanese person is directly to right of the person in the studio.
:- person_is(H1, japanese), house_is(H2, studio), H1 != H2+1.

% Clue 17: The person in the farmhouse is adjacent to the person whose favorite fictional genre is sci-fi.
:- house_is(H1, farmhouse), likes(H2, sci_fi), |H1-H2| != 1.

% Clue 18: There are two houses in between the person who smokes marlboro and the person who wears the t-shirt.
:- smokes(H1, marlboro), wears(H2, t_shirt), |H1-H2| != 3.

% Clue 19: The person whose favorite fictional genre is satire is somewhere to the left of the person who smokes camel.
:- likes(H1, satire), smokes(H2, camel), H1 >= H2.

% Clue 20: The person who smokes l&m is not the person who wears the belt.
:- smokes(H, l_m), wears(H, belt).

% Clue 21: The astronaut is directly to left of the person in the apartment.
:- works_as(H1, astronaut), house_is(H2, apartment), H1 != H2-1.

% Clue 22: The person who wears the jeans is in house number 4.
:- not wears(4, jeans).

% Clue 23: The person in the townhouse is somewhere to the right of the person who wears the belt.
:- house_is(H1, townhouse), wears(H2, belt), H1 <= H2.

% Clue 24: There is one house in between the person in the loft and the indian person.
:- house_is(H1, loft), person_is(H2, indian), |H1-H2| != 2.

% Clue 25: The person who wears the jeans is adjacent to the person who wears the watch.
:- wears(H1, jeans), wears(H2, watch), |H1-H2| != 1.
% module constraints END

% module display
% Show the assignments for each category to verify the solution.
#show house_is/2.
#show person_is/2.
#show smokes/2.
#show likes/2.
#show wears/2.
#show works_as/2.
% module display END
% MODULES END
```
\end{lstlisting}

\subsubsection{{\sc Clingo} Feedback}

This program is automatically run with {\sc clingo}, which produces the following answer set.

\begin{lstlisting}[escapechar=^]
Solving...

Answer Set 1: 

wears(4,jeans) wears(3,watch) house_is(3,loft) person_is(5,indian) wears(1,belt) house_is(6,townhouse) works_as(2,astronaut) house_is(3,apartment) smokes(3,l_m) likes(1,satire) smokes(6,camel) smokes(6,marlboro) wears(3,t_shirt) likes(1,sci_fi) house_is(2,farmhouse) person_is(4,japanese) house_is(3,studio) person_is(3,norwegian) person_is(5,chinese) works_as(5,actor) person_is(4,italian) likes(4,dystopian) person_is(1,mexican) wears(2,vest) works_as(1,architect) likes(2,fairy_tale) smokes(1,benson_hedges) likes(5,war) works_as(1,lawyer) house_is(3,cabin) works_as(3,mechanic) smokes(5,lucky_strike) smokes(1,prince) works_as(2,magician) likes(2,drama) wears(3,coat)

SATISFIABLE
\end{lstlisting}

\subsection{2nd Pass}

In the second pass, the LLM generator receives the previously generated program and answer set produced.

\subsubsection{Input} 

\begin{lstlisting}[escapechar=^]
Consider the following resource material describing Answer Set Programming and the language of Clingo to do the following task.
<resource-material>
^{\cblu \textlangle resource omitted\textrangle}^ 
</resource-material>


<instructions>
You are tasked with constructing an ASP program to solve the following problem:
<problem-description>
^{\cblu \textlangle input problem omitted, see above\textrangle}^ 
IMPORTANT: Atoms with large arity may take very long for larger problems, so do not introduce them unnecessarily.
</problem-description>
The ASP program should be made of ASP modules, which each make up a necessary part of the program to represent the problem.
In the program, include the #show command for atoms which will help for verifying the answer and/or debugging. For example, don't just show the final answer, but show atoms which can be used to check that the parts determining the final answer are also correct.

<current-clingo-modules>
^{\cblu \textlangle previously generated ASP program omitted, see above\textrangle}^ 
</current-clingo-modules>
<candidate-answer-set>
Here is the candidate answer set to solve the problem, from running the ASP program:
stdout:
Solving...

Answer Set 1: 

wears(4,jeans) wears(3,watch) house_is(3,loft) person_is(5,indian) wears(1,belt) house_is(6,townhouse) works_as(2,astronaut) house_is(3,apartment) smokes(3,l_m) likes(1,satire) smokes(6,camel) smokes(6,marlboro) wears(3,t_shirt) likes(1,sci_fi) house_is(2,farmhouse) person_is(4,japanese) house_is(3,studio) person_is(3,norwegian) person_is(5,chinese) works_as(5,actor) person_is(4,italian) likes(4,dystopian) person_is(1,mexican) wears(2,vest) works_as(1,architect) likes(2,fairy_tale) smokes(1,benson_hedges) likes(5,war) works_as(1,lawyer) house_is(3,cabin) works_as(3,mechanic) smokes(5,lucky_strike) smokes(1,prince) works_as(2,magician) likes(2,drama) wears(3,coat)

SATISFIABLE



Models       : 1+


</candidate-answer-set>


You are tasked with doing an operation from the following.
(UPDATE) - This should be done either to:
	(a) write an initial ASP program (still considered an update).
	(b) update the program based on the Clingo output which may help to debug the current output.
If updated, make sure to write the complete program.
(PASS) - This should be done when the Clingo output is correct. There should be no ambiguity and consensus on the proposed solution being correct. ONLY use when absolutely sure the output is correct. If the proposed solution is correct, you may ignore stderr and still pass.

For UPDATE, format your output exactly like the following, writing the complete program:

```
% MODULES START
% module <name of first module>
<ASP code for this module>
% module <name of first module> END

% module <name of second module>
<ASP code for this module>
% module <name of second module> END
...
% MODULES END
```
For PASS, format your output exactly like the following (only write the action, since nothing else is to be done):

```
OPERATION: PASS
```

Do not write anything outside of the three backticks.
</instructions>
\end{lstlisting}

\subsubsection{Output}

Based on the instruction from the previous prompt, if UPDATE is chosen, then the LLM should simply rewrite the program. Otherwise, it should write PASS in its output. The LLM chooses not to pass the resulting answer set from the previously generated ASP program by supplying an updated program. This is likely due to the apparent issue with the answer set; it assigns multiple attributes from the same category to a single house (e.g., the person in house 1 is both an architect and a lawyer).

While the program specifies that each attribute within a category must be assigned to a house with the following:

\texttt{\{ works\_as(H, O) : house(H) \} = 1 :- occupation(O).}

However, it does not enforce that every house must receive exactly one attribute from each category:

\texttt{\{ works\_as(H, O) : occupation(O) \} = 1 :- house(H).}
The program is updated as follows.

\begin{lstlisting}[escapechar=^]
```
% MODULES START
% module domain
% Houses
house(1..6).

% Categories and their unique attributes
house_type(farmhouse; cabin; studio; apartment; loft; townhouse).
nationality(norwegian; japanese; italian; chinese; mexican; indian).
cigarette(prince; camel; lucky_strike; benson_hedges; marlboro; l_m).
genre(satire; sci_fi; dystopian; fairy_tale; drama; war).
clothing(vest; belt; jeans; t_shirt; coat; watch).
occupation(lawyer; actor; astronaut; magician; mechanic; architect).
% module domain END

% module generate
% Generate a unique assignment for each attribute to a house.
% For each category, this creates a one-to-one mapping (a bijection)
% between the attributes of that category and the houses.

% house_is(House, Type)
{ house_is(H, T) : house_type(T) } = 1 :- house(H).
{ house_is(H, T) : house(H) } = 1 :- house_type(T).

% person_is(House, Nationality)
{ person_is(H, N) : nationality(N) } = 1 :- house(H).
{ person_is(H, N) : house(H) } = 1 :- nationality(N).

% smokes(House, Brand)
{ smokes(H, C) : cigarette(C) } = 1 :- house(H).
{ smokes(H, C) : house(H) } = 1 :- cigarette(C).

% likes(House, Genre)
{ likes(H, G) : genre(G) } = 1 :- house(H).
{ likes(H, G) : house(H) } = 1 :- genre(G).

% wears(House, Item)
{ wears(H, I) : clothing(I) } = 1 :- house(H).
{ wears(H, I) : house(H) } = 1 :- clothing(I).

% works_as(House, Occupation)
{ works_as(H, O) : occupation(O) } = 1 :- house(H).
{ works_as(H, O) : house(H) } = 1 :- occupation(O).
% module generate END

% module constraints
% This module translates the clues into integrity constraints.

% Clue 1: There are two houses in between the person who smokes lucky strike and the person whose favorite fictional genre is drama.
:- smokes(H1, lucky_strike), likes(H2, drama), |H1-H2| != 3.

% Clue 2: The person who smokes prince is directly to left of the magician.
:- smokes(H1, prince), works_as(H2, magician), H1 != H2-1.

% Clue 3: There are two houses in between the person who smokes lucky strike and the person who wears the vest.
:- smokes(H1, lucky_strike), wears(H2, vest), |H1-H2| != 3.

% Clue 4: The person in the cabin and the mechanic are the same person.
:- house_is(H1, cabin), works_as(H2, mechanic), H1 != H2.

% Clue 5: The lawyer is in house number 1.
:- not works_as(1, lawyer).

% Clue 6: The indian person and the person whose favorite fictional genre is war are the same person.
:- person_is(H1, indian), likes(H2, war), H1 != H2.

% Clue 7: There are two houses in between the person who smokes benson & hedges and the person whose favorite fictional genre is dystopian.
:- smokes(H1, benson_hedges), likes(H2, dystopian), |H1-H2| != 3.

% Clue 8: The person whose favorite fictional genre is satire is adjacent to the person whose favorite fictional genre is fairy tale.
:- likes(H1, satire), likes(H2, fairy_tale), |H1-H2| != 1.

% Clue 9: The person whose favorite fictional genre is fairy tale is somewhere to the right of the architect.
:- likes(H1, fairy_tale), works_as(H2, architect), H1 <= H2.

% Clue 10: The person who wears the vest is somewhere to the right of the person whose favorite fictional genre is sci-fi.
:- wears(H1, vest), likes(H2, sci_fi), H1 <= H2.

% Clue 11: The mexican person is in house number 1.
:- not person_is(1, mexican).

% Clue 12: The italian person and the person whose favorite fictional genre is dystopian are the same person.
:- person_is(H1, italian), likes(H2, dystopian), H1 != H2.

% Clue 13: The person in the farmhouse is directly to left of the person in the apartment.
:- house_is(H1, farmhouse), house_is(H2, apartment), H1 != H2-1.

% Clue 14: The chinese person and the actor are the same person.
:- person_is(H1, chinese), works_as(H2, actor), H1 != H2.

% Clue 15: The norwegian person is adjacent to the astronaut.
:- person_is(H1, norwegian), works_as(H2, astronaut), |H1-H2| != 1.

% Clue 16: The japanese person is directly to right of the person in the studio.
:- person_is(H1, japanese), house_is(H2, studio), H1 != H2+1.

% Clue 17: The person in the farmhouse is adjacent to the person whose favorite fictional genre is sci-fi.
:- house_is(H1, farmhouse), likes(H2, sci_fi), |H1-H2| != 1.

% Clue 18: There are two houses in between the person who smokes marlboro and the person who wears the t-shirt.
:- smokes(H1, marlboro), wears(H2, t_shirt), |H1-H2| != 3.

% Clue 19: The person whose favorite fictional genre is satire is somewhere to the left of the person who smokes camel.
:- likes(H1, satire), smokes(H2, camel), H1 >= H2.

% Clue 20: The person who smokes l&m is not the person who wears the belt.
:- smokes(H, l_m), wears(H, belt).

% Clue 21: The astronaut is directly to left of the person in the apartment.
:- works_as(H1, astronaut), house_is(H2, apartment), H1 != H2-1.

% Clue 22: The person who wears the jeans is in house number 4.
:- not wears(4, jeans).

% Clue 23: The person in the townhouse is somewhere to the right of the person who wears the belt.
:- house_is(H1, townhouse), wears(H2, belt), H1 <= H2.

% Clue 24: There is one house in between the person in the loft and the indian person.
:- house_is(H1, loft), person_is(H2, indian), |H1-H2| != 2.

% Clue 25: The person who wears the jeans is adjacent to the person who wears the watch.
:- wears(H1, jeans), wears(H2, watch), |H1-H2| != 1.
% module constraints END

% module display
% Show the assignments for each category to verify the solution.
#show house_is/2.
#show person_is/2.
#show smokes/2.
#show likes/2.
#show wears/2.
#show works_as/2.
% module display END
% MODULES END
```
\end{lstlisting}

In this case, the LLM generator decides to update the program, adding the following missing rules to the ASP program, in the generate module.
\begin{lstlisting}[escapechar=^]
% house_is(House, Type)
{ house_is(H, T) : house(H) } = 1 :- house_type(T).
{ person_is(H, N) : house(H) } = 1 :- nationality(N).
{ smokes(H, C) : house(H) } = 1 :- cigarette(C).
{ likes(H, G) : house(H) } = 1 :- genre(G).
{ wears(H, I) : house(H) } = 1 :- clothing(I).
{ works_as(H, O) : house(H) } = 1 :- occupation(O).
\end{lstlisting}

\subsubsection{{\sc Clingo} Feedback}
The associated answer set produced now represents the correct assignment.
\begin{lstlisting}[escapechar=^]
Solving...

Answer Set 1: 

wears(4,jeans) wears(5,watch) house_is(4,loft) person_is(6,indian) wears(3,belt) house_is(5,townhouse) works_as(2,astronaut) house_is(3,apartment) smokes(1,l_m) likes(3,satire) smokes(6,camel) smokes(3,marlboro) wears(6,t_shirt) likes(1,sci_fi) house_is(2,farmhouse) house_is(1,studio) person_is(2,japanese) person_is(3,norwegian) person_is(4,chinese) works_as(4,actor) person_is(5,italian) likes(5,dystopian) person_is(1,mexican) wears(2,vest) works_as(3,architect) likes(4,fairy_tale) smokes(2,benson_hedges) likes(6,war) works_as(1,lawyer) house_is(6,cabin) works_as(6,mechanic) smokes(5,lucky_strike) smokes(4,prince) works_as(5,magician) likes(2,drama) wears(1,coat)

SATISFIABLE
\end{lstlisting}

\subsection{3rd Pass}
Again, the LLM generator views the previously updated program and the (correct) answer set, and is tasked with either updating or passing the program/output.

\subsubsection{Input}

\begin{lstlisting}[escapechar=^]
Consider the following resource material describing Answer Set Programming and the language of Clingo to do the following task.
<resource-material>
^{\cblu \textlangle resource omitted\textrangle}^ 
</resource-material>


<instructions>
You are tasked with constructing an ASP program to solve the following problem:
<problem-description>
^{\cblu \textlangle input problem description omitted, see above\textrangle}^ 
IMPORTANT: Atoms with large arity may take very long for larger problems, so do not introduce them unnecessarily.
</problem-description>
The ASP program should be made of ASP modules, which each make up a necessary part of the program to represent the problem.
In the program, include the #show command for atoms which will help for verifying the answer and/or debugging. For example, don't just show the final answer, but show atoms which can be used to check that the parts determining the final answer are also correct.

<current-clingo-modules>
^{\cblu \textlangle previously generated ASP program omitted, see above \textrangle}^ 
</current-clingo-modules>
<candidate-answer-set>
Here is the candidate answer set to solve the problem, from running the ASP program:
stdout:
Solving...

Answer Set 1: 

wears(4,jeans) wears(5,watch) house_is(4,loft) person_is(6,indian) wears(3,belt) house_is(5,townhouse) works_as(2,astronaut) house_is(3,apartment) smokes(1,l_m) likes(3,satire) smokes(6,camel) smokes(3,marlboro) wears(6,t_shirt) likes(1,sci_fi) house_is(2,farmhouse) house_is(1,studio) person_is(2,japanese) person_is(3,norwegian) person_is(4,chinese) works_as(4,actor) person_is(5,italian) likes(5,dystopian) person_is(1,mexican) wears(2,vest) works_as(3,architect) likes(4,fairy_tale) smokes(2,benson_hedges) likes(6,war) works_as(1,lawyer) house_is(6,cabin) works_as(6,mechanic) smokes(5,lucky_strike) smokes(4,prince) works_as(5,magician) likes(2,drama) wears(1,coat)

SATISFIABLE



Models       : 1+


</candidate-answer-set>


You are tasked with doing an operation from the following.
(UPDATE) - This should be done either to:
	(a) write an initial ASP program (still considered an update).
	(b) update the program based on the Clingo output which may help to debug the current output.
If updated, make sure to write the complete program.
(PASS) - This should be done when the Clingo output is correct. There should be no ambiguity and consensus on the proposed solution being correct. ONLY use when absolutely sure the output is correct. If the proposed solution is correct, you may ignore stderr and still pass.

For UPDATE, format your output exactly like the following, writing the complete program:

```
% MODULES START
% module <name of first module>
<ASP code for this module>
% module <name of first module> END

% module <name of second module>
<ASP code for this module>
% module <name of second module> END
...
% MODULES END
```
For PASS, format your output exactly like the following (only write the action, since nothing else is to be done):

```
OPERATION: PASS
```

Do not write anything outside of the three backticks.
</instructions>
\end{lstlisting}

\subsubsection{Output}

The LLM generator passes the program.

\begin{lstlisting}[escapechar=^]
```
OPERATION: PASS
```
\end{lstlisting}

\subsection{Human readable formatting}

After passing, an LLM is now tasked with generating a human-readable version of the answer set as the final solution.

\subsubsection{Input}

\begin{lstlisting}[escapechar=^]
You are tasked with writing the answer set produced by running Clingo into a human-readable form.

Here is the problem description:
^{\cblu \textlangle input problem description omitted, see above \textrangle}^ 

Clingo output:
stdout:
Solving...

Answer Set 1: 

wears(4,jeans) wears(5,watch) house_is(4,loft) person_is(6,indian) wears(3,belt) house_is(5,townhouse) works_as(2,astronaut) house_is(3,apartment) smokes(1,l_m) likes(3,satire) smokes(6,camel) smokes(3,marlboro) wears(6,t_shirt) likes(1,sci_fi) house_is(2,farmhouse) house_is(1,studio) person_is(2,japanese) person_is(3,norwegian) person_is(4,chinese) works_as(4,actor) person_is(5,italian) likes(5,dystopian) person_is(1,mexican) wears(2,vest) works_as(3,architect) likes(4,fairy_tale) smokes(2,benson_hedges) likes(6,war) works_as(1,lawyer) house_is(6,cabin) works_as(6,mechanic) smokes(5,lucky_strike) smokes(4,prince) works_as(5,magician) likes(2,drama) wears(1,coat)

SATISFIABLE



Models       : 1+



Your conversion should be encapsulated completely in 3 backticks, without any extraneous text. It should be in natural language, and precise. If the Clingo output is unsatisfiable, or the program produces an error, the write "UNSAT" or "ERROR" encapsulated in 3 backticks.
\end{lstlisting}

\subsubsection{Output}
A human-readable solution is generated.
\begin{lstlisting}[escapechar=^]
- **House 1**: The Mexican lawyer lives in the studio, likes sci-fi, smokes L&M, and wears a coat.
- **House 2**: The Japanese astronaut lives in the farmhouse, likes drama, smokes Benson & Hedges, and wears a vest.
- **House 3**: The Norwegian architect lives in the apartment, likes satire, smokes Marlboro, and wears a belt.
- **House 4**: The Chinese actor lives in the loft, likes fairy tale, smokes Prince, and wears jeans.
- **House 5**: The Italian magician lives in the townhouse, likes dystopian, smokes Lucky Strike, and wears a watch.
- **House 6**: The Indian mechanic lives in the cabin, likes war, smokes Camel, and wears a t-shirt.
\end{lstlisting}

\section{Expanded Reference Effect Analysis} \label{appendix:error-analysis-zl-xxl}

On \textbf{ZL-XXL}, we measure the number of occurrences for common issues among the reasoning models.

We observe that with the standard reference, {\sf LLM+ASP} more often fails to write all constraints of the problem. When using the standard reference, { the number of programs with missing constraints} for \textbf{ZL-XXL} increases by 43\% over without any reference (from 46 to 66) as shown in Table~\ref{tab:llm-judge-issues}. When using the compact reference, this decreases by 30.4\% (from 46 to 32).
Since missing constraints are not related to the content in the reference, this suggests that a longer prompt makes it harder for the LLM to be as comprehensive as it should be, failing to incorporate important details in the input problem. 

Moreover, in our experiments, the LLMs do not use the reference as well when it is longer. For example, both the standard reference and compact reference have the same content regarding pooling and operator use, but the standard reference is about 8 times the length. Despite this, with the standard reference {\sf LLM+ASP} more frequently produces incorrect syntax for both, as reported in Table~\ref{tab:llm-judge-issues}. As compared without a reference, operator and pooling issues decrease by 82.5\% and 25\% respectively with the standard reference. However, with the compact reference, this is reduced further, by 92.5\% and 70.5\%. Prior studies that show degradation of LLM performance on both retrieval and reasoning when introducing extra context are consistent with our work \cite{kuratov24babilong, chatziveroglou25exploring, hengle25can, hong25context}. These works show that the length of the context both with and without distractions affects performance, and that LLMs only use a fraction of their full available context effectively. With the standard reference and additional text from problem descriptions, written programs, and {\sc clingo} feedback, the prompt size approaches 30k tokens, which is past where performance starts to degrade.

For the planning domain Mystery Blocksworld, without a reference, the LLMs never used incremental {\sc Clingo} where useful. Despite a description and example of it in the standard reference, incremental {\sc Clingo} is only used 22\% of the time with o4-mini, and 0\% of the time with DeepSeek R1. 
With the compact reference it is used 76\%, and 94\% of the time, respectively.
{ 
These observations suggest that the effectiveness of any in-prompt reference is partly constrained by what the LLM has already seen during pre-training. This points to a complementary path for improving {\sf LLM+ASP}: fine-tuning LLMs on targeted ASP materials (e.g., operator semantics, pooling patterns, incremental {\sc clingo} for actions and effects) to internalize these conventions rather than relying solely on external reference prompts.
}

\begin{table}[h]
\footnotesize
    \caption{Categorized issues for {\sf LLM+ASP} on the \textbf{ZL-XXL} benchmark.}
    \centering
    \begin{tabular}{l|c|c|c|c|c}
        \toprule
        Mode & \textbf{Flash} & \textbf{Pro} & \textbf{o4-mini} & \textbf{R1} & \textbf{Total}\\ 
        \midrule
\multicolumn{6}{c}{\textbf{LLM + ASP Without Reference}} \\
\midrule
        Operator & 36 & 0 & 4 & 0 & 40\\ \hline
        Pooling & 20 & 11 & 7 & 6 & 44 \\ \hline
        Missing Cons. & 14 & 3 & 19 & 10 & 46 \\ \hline
        Timeout & 0 & 0 & 2 & 5 & 7\\ \hline
        Total & 70 & 14 & 32 & 21 & 137\\
\midrule
\multicolumn{6}{c}{\textbf{LLM + ASP With {Standard} Reference}} \\
\midrule

        Operator & 6 & 0 & 1 & 0 & 7\\ \hline
        Pooling & 18 & 5 & 6 & 4 & 33\\ \hline
        Missing Cons. & 31 & 0 & 19 & 16 & 66 \\ \hline
        Timeout & 0 & 0 & 0 & 4 & 4\\ \hline
        Total & 55 & 5 & 26 & 24 & 110 \\ \hline

\multicolumn{6}{c}{\textbf{LLM + ASP With Compact Reference}} \\
\midrule
        Operator & 2 & 0 & 1 & 0 & 3\\ \hline
        Pooling & 7 & 1 & 2 & 3 & 13\\ \hline
        Missing Cons. & 10 & 0 & 14 & 8 & 32\\ \hline
        Timeout & 0 & 0 & 1 & 7 & 8\\ \hline
        Total & 19 & 1 & 18 & 18 & 56 \\ 
\midrule
    \end{tabular}
    \label{tab:llm-judge-issues}
\end{table}

\section{Expanded Analysis of Iterative Corrections \label{appendix:iterative-corrections-no-reference}}

In Table~\ref{tab:revision-comparison-no-reference} we present the performance of {\sf LLM+ASP} without any reference, and with and without any revisions.
\begin{table*}[ht]
    \centering
    \caption{{\sf LLM+ASP} Performance with no reference (comparing without revisions and with revisions).}
    \footnotesize
    \setlength{\tabcolsep}{4pt}
    \label{tab:revision-comparison-no-reference}
    \begin{tabular}{lcccccccccccc}
        \toprule
        \multirow{2}{*}{\textbf{LLM Model}} & \multicolumn{3}{c}{\textbf{ZL-XL (\%)}} & \multicolumn{3}{c}{\textbf{ZL-XXL (\%)}} & \multicolumn{3}{c}{\textbf{SudokuBench (\%)}} & \multicolumn{3}{c}{\textbf{MysteryBW (\%)}} \\
        \cmidrule(lr){2-4} \cmidrule(lr){5-7} \cmidrule(lr){8-10} \cmidrule(lr){11-13}
        & \subhead & \subheadrev & \subheadavg & \subhead & \subheadrev & \subheadavg & \subhead & \subheadrev & \subheadavg & \subhead & \subheadrev & \subheadavg \\
        \midrule
        Gemini 2.5 Pro & 88.0 & 97.0 & 0.15 & 68.0 & 88.0 & 2.33 & 71.0 & 76.0 & 1.96 & 48.0 & 69.0 & 1.12 \\
        Gemini 2.5 Flash & 15.0 & 38.0 & 1.49 & 2.0 & 25.0 & 2.80 & 35.0 & 53.0 & 2.94 & 23.0 & 52.0 & 1.29 \\
        o4-mini & 58.0 & 97.0 & 0.84 & 29.0 & 67.0 & 3.29 & 21.0 & 72.0 & 3.29 & 5.0 & 41.0 & 4.49 \\
        DS-R1-0528 & 64.0 & 95.0 & 0.69 & 47.0 & 76.0 & 1.90 & 28.0 & 68.0 & 3.76 & 24.0 & 64.0 & 1.86 \\
        DS-V3 & 18.0 & 49.0 & 4.90 & 11.0 & 28.0 & 7.67 & 12.0 & 22.0 & 7.68 & 3.0 & 26.0 & 7.77 \\
        \textbf{Average} & \textbf{48.6} & \textbf{75.2} & \textbf{1.61} & \textbf{31.4} & \textbf{56.8} & \textbf{3.60} & \textbf{33.4} & \textbf{58.2} & \textbf{3.93} & \textbf{20.6} & \textbf{50.4} & \textbf{3.31} \\
        
        \midrule \midrule
        
        \multirow{2}{*}{\textbf{LLM Model}} & \multicolumn{3}{c}{\textbf{Skeptical}} & \multicolumn{3}{c}{\textbf{Credulous}} & \multicolumn{3}{c}{\textbf{BoardgameQA}} & \multicolumn{3}{c}{\textbf{Overall Average}} \\
        \cmidrule(lr){2-4} \cmidrule(lr){5-7} \cmidrule(lr){8-10} \cmidrule(lr){11-13}
        & \subhead & \subheadrev & \subheadavg & \subhead & \subheadrev & \subheadavg & \subhead & \subheadrev & \subheadavg & \subhead & \subheadrev & \subheadavg \\
        \midrule
        Gemini 2.5 Pro    & 98.5 & 100.0 & 0.0 & 96.0 & 96.5 & 0.2 & 67.5 & 96.0 & 1.3 & 76.7 & 87.4 & 1.0 \\
        Gemini 2.5 Flash  & 96.5 & 99.5 & 0.1 & 97.5 & 98.5 & 0.1 & 67.5 & 92.5 & 1.8 & 48.1 & 65.5 & 1.5 \\
        o4-mini           & 93.0 & 96.5 & 0.5 & 89.5 & 82.0 & 0.8 & 71.0 & 91.0 & 0.9 & 52.4 & 79.4 & 2.0 \\
        DS-R1-0528           & 86.5 & 91.5 & 1.4 & 89.0 & 87.5 & 1.0 & 74.0 & 89.5 & 2.6 & 58.9 & 81.6 & 1.9 \\
        DS-V3       & 68.5 & 77.0 & 3.4 & 65.0 & 57.0 & 3.9 & 37.0 & 41.5 & 6.5 & 30.6 & 42.9 & 6.0 \\
        
        \textbf{Average} & \textbf{88.6} & \textbf{92.9} & \textbf{1.1} & \textbf{87.4} & \textbf{84.3} & \textbf{1.2} & \textbf{63.4} & \textbf{82.1} & \textbf{2.6} & \textbf{53.3} & \textbf{71.4} & \textbf{2.5} \\
        \bottomrule
    \end{tabular}
\end{table*}

\section{Expanded Analysis of Failure Modes of {\sf LLM+ASP}} \label{appendix:expanded-tables}

This section expands on the failure mode analysis from Table~\ref{tab:errors-with-reference} in the main body, breaking down the errors by model for each dataset. The detailed statistics are presented across the following four tables, categorized by whether an external reference guide was used and whether iterative revisions were permitted:

\begin{itemize}
\item With Standard Reference:

\begin{itemize}
\item No Revisions: Table~\ref{tab:error_cases-without-revisions-with-reference}

\item With Revisions: Table~\ref{tab:error_cases-with-revisions-with-reference}
\end{itemize}

\item Without Reference:

\begin{itemize}
\item 
No Revisions: Table~\ref{tab:error_cases-without-revisions-without-reference}

\item With Revisions: Table~\ref{tab:error_cases-with-revisions-without-reference}
\end{itemize}
\end{itemize}

From this detailed breakdown, we can observe that using an external reference guide helps reduce the incidence of programs that are unsatisfiable and shifts the error profile toward cases that are satisfiable-but-incorrect.

When revisions are permitted, the percentage of errors that are unsatisfiable drops from 66.96\% (without a reference) to 40.3\% (with a reference). Likewise, the proportion of satisfiable-but-incorrect programs increases from 28.12\% to 49.2\%. This reflects a shift toward subtler semantic errors, which, because they produce a concrete answer set, may be more correctable than  unsatisfiable results that provide less feedback.

\begin{table*}[b]
\centering
\caption{Error cases with no revisions and with a reference.}
\label{tab:error_cases-without-revisions-with-reference}
\footnotesize 
\begin{tabular}{@{}llrrrrr@{}}
\toprule
\textbf{Dataset} & \textbf{Model} & \textbf{Syntax} & \textbf{Timeout} & \textbf{Sat} & \textbf{Unsat} & \textbf{Total} \\
\midrule
\multicolumn{7}{l}{\textbf{ZL-XL}} \\
& Gemini 2.5 Pro + ASP & 3 & 0 & 15 & 7 & 25 \\
& DS-R1-0528 + ASP & 8 & 0 & 4 & 5 & 17 \\
& o4-mini + ASP & 37 & 0 & 2 & 4 & 43 \\
& Gemini 2.5 Flash + ASP & 31 & 0 & 6 & 7 & 44 \\
& DS-V3 + ASP & 23 & 0 & 17 & 19 & 59 \\
\cmidrule(l){2-7}
& \textbf{Total} & \textbf{102} & \textbf{0} & \textbf{44} & \textbf{42} & \textbf{188} \\
& \textit{Percentage} & \textit{54.26\%} & \textit{0.00\%} & \textit{23.40\%} & \textit{22.34\%} & \\
& \textit{Percentage excluding DS-V3} &\textit{ 61.2\%} & \textit{0.0\%} & \textit{20.9\%} & \textit{17.8\%} \\
\midrule
\multicolumn{7}{l}{\textbf{ZL-XXL}} \\
& Gemini 2.5 Pro + ASP & 1 & 4 & 11 & 15 & 31 \\
& DS-R1-0528 + ASP & 16 & 10 & 21 & 5 & 52 \\
& o4-mini + ASP & 62 & 1 & 7 & 4 & 74 \\
& Gemini 2.5 Flash + ASP & 44 & 0 & 15 & 18 & 77 \\
& DS-V3 + ASP & 58 & 1 & 14 & 16 & 89 \\
\cmidrule(l){2-7}
& \textbf{Total} & \textbf{181} & \textbf{16} & \textbf{68} & \textbf{58} & \textbf{323} \\
& \textit{Percentage} & \textit{56.04\%} & \textit{4.95\%} & \textit{21.05\%} & \textit{17.96\%} & \\
& \textit{Percentage excluding DS-V3} &  \textit{52.6\%} & \textit{6.4\%} & \textit{23.1\% }& \textit{18.0 \%} \\
\midrule
\multicolumn{7}{l}{\textbf{SudokuBench}} \\
& Gemini 2.5 Pro + ASP & 14 & 1 & 13 & 16 & 44 \\
& DS-R1-0528 + ASP & 37 & 1 & 13 & 17 & 68 \\
& o4-mini + ASP & 63 & 0 & 3 & 7 & 73 \\
& Gemini 2.5 Flash + ASP & 44 & 0 & 7 & 11 & 62 \\
& DS-V3 + ASP & 73 & 0 & 9 & 8 & 90 \\
\cmidrule(l){2-7}
& \textbf{Total} & \textbf{231} & \textbf{2} & \textbf{45} & \textbf{59} & \textbf{337} \\
& \textit{Percentage} & \textit{68.55\%} & \textit{0.59\%} & \textit{13.35\%} & \textit{17.51\%} & \\
& \textit{Percentage excluding DS-V3} & \textit{64.0 \%} & \textit{0.8\%} & \textit{14.6\%} & \textit{20.7\%} \\
\midrule[\heavyrulewidth]
\multicolumn{2}{l}{\textbf{Total}} & \textbf{514} & \textbf{18} & \textbf{157} & \textbf{159} & \textbf{848} \\
\multicolumn{2}{l}{\textbf{Total excluding DS-V3}} & \textbf{360} & \textbf{17} & \textbf{117} & \textbf{116} & \textbf{610} \\
\multicolumn{2}{l}{\textit{\textbf{Total Percentage}}} & \textit{\textbf{60.61\%}} & \textit{\textbf{2.12\%}} & \textit{\textbf{18.51\%}} & \textit{\textbf{18.75\%}} & \\
\multicolumn{2}{l}{\textit{\textbf{Total Percentage excluding DS-V3}}} & \textit{59.0\%} & \textit{2.8\%} & \textit{19.2\%} & \textit{19.0\%} \\
\bottomrule
\end{tabular}
\end{table*}

\begin{table*}[b]
\centering
\caption{Error cases with revisions and a reference.}
\label{tab:error_cases-with-revisions-with-reference}
\small 
\begin{tabular}{@{}llrrrrr@{}}
\toprule
\textbf{Dataset} & \textbf{Model} & \textbf{Syntax} & \textbf{Timeout} & \textbf{Sat} & \textbf{Unsat} & \textbf{Total} \\
\midrule
\multicolumn{7}{l}{\textbf{ZL-XL}} \\
& Gemini 2.5 Pro + ASP & 0 & 0 & 0 & 5 & 5 \\
& DS-R1-0528 + ASP & 0 & 0 & 2 & 0 & 2 \\
& o4-mini + ASP & 0 & 0 & 1 & 1 & 2 \\
& Gemini 2.5 Flash + ASP & 0 & 0 & 1 & 10 & 11 \\
& DS-V3 + ASP & 1 & 0 & 15 & 21 & 37 \\
\cmidrule(l){2-7}
& \textbf{Total} & \textbf{1} & \textbf{0} & \textbf{19} & \textbf{37} & \textbf{57} \\
& \textit{Percentage} & \textit{1.75\%} & \textit{0.00\%} & \textit{33.33\%} & \textit{64.91\%} & \\
& \textit{Percentage excluding DS-V3} & \textit{0.00\%} & \textit{0.00\%} & \textit{20.00\%} & \textit{80.00\%} & \\
\midrule
\multicolumn{7}{l}{\textbf{ZL-XXL}} \\
& Gemini 2.5 Pro + ASP & 0 & 0 & 2 & 4 & 6 \\
& DS-R1-0528 + ASP & 6 & 4 & 17 & 5 & 32 \\
& o4-mini + ASP & 1 & 0 & 24 & 7 & 32 \\
& Gemini 2.5 Flash + ASP & 0 & 0 & 15 & 31 & 46 \\
& DS-V3 + ASP & 20 & 0 & 14 & 42 & 76 \\
\cmidrule(l){2-7}
& \textbf{Total} & \textbf{27} & \textbf{4} & \textbf{72} & \textbf{89} & \textbf{192} \\
& \textit{Percentage} & \textit{14.06\%} & \textit{2.08\%} & \textit{37.50\%} & \textit{46.35\%} & \\
& \textit{Percentage excluding DS-V3} & \textit{6.03\%} & \textit{3.45\%} & \textit{50.00\%} & \textit{40.52\%} & \\
\midrule
\multicolumn{7}{l}{\textbf{SudokuBench}} \\
& Gemini 2.5 Pro + ASP & 1 & 2 & 16 & 6 & 25 \\
& DS-R1-0528 + ASP & 4 & 1 & 15 & 12 & 32 \\
& o4-mini + ASP & 4 & 1 & 16 & 14 & 35 \\
& Gemini 2.5 Flash + ASP & 2 & 1 & 18 & 9 & 30 \\
& DS-V3 + ASP & 32 & 26 & 16 & 0 & 74 \\
\cmidrule(l){2-7}
& \textbf{Total} & \textbf{43} & \textbf{31} & \textbf{81} & \textbf{41} & \textbf{196} \\
& \textit{Percentage} & \textit{21.94\%} & \textit{15.82\%} & \textit{41.33\%} & \textit{20.92\%} & \\
& \textit{Percentage excluding DS-V3} & \textit{9.02\%} & \textit{4.10\%} & \textit{53.28\%} & \textit{33.61\%} & \\
\midrule[\heavyrulewidth]
\multicolumn{2}{l}{\textbf{Total}} & \textbf{71} & \textbf{35} & \textbf{172} & \textbf{167} & \textbf{445} \\
\multicolumn{2}{l}{\textbf{Total excluding DS-V3}} & \textbf{11} & \textbf{5} & \textbf{65} & \textbf{41} & \textbf{122} \\
\multicolumn{2}{l}{\textit{\textbf{Total Percentage}}} & \textit{\textbf{15.96\%}} & \textit{\textbf{7.87\%}} & \textit{\textbf{38.65\%}} & \textit{\textbf{37.53\%}} & \\
\multicolumn{2}{l}{\textit{\textbf{Total Percentage excluding DS-V3}}} & \textit{\textbf{6.98\%}} & \textit{\textbf{3.49\%}} & \textit{\textbf{49.22\%}} & \textit{\textbf{40.31\%}} & \\
\bottomrule
\end{tabular}
\end{table*}

\begin{table*}[htbp]
\centering
\caption{Error cases with no revisions and without a reference.}
\label{tab:error_cases-without-revisions-without-reference}
\footnotesize 
\begin{tabular}{@{}llrrrrr@{}}
\toprule
\textbf{Dataset} & \textbf{Model} & \textbf{Syntax} & \textbf{Timeout} & \textbf{Sat} & \textbf{Unsat} & \textbf{Total} \\
\midrule
\multicolumn{7}{l}{\textbf{ZL-XL}} \\
& Gemini 2.5 Pro + ASP & 2 & 0 & 5 & 4 & 11 \\
& DS-R1-0528 + ASP & 17 & 0 & 16 & 3 & 36 \\
& o4-mini + ASP & 25 & 0 & 2 & 14 & 41 \\
& Gemini 2.5 Flash + ASP & 43 & 0 & 24 & 18 & 85 \\
& DS-V3 + ASP & 41 & 0 & 25 & 16 & 82 \\
\cmidrule(l){2-7}
& \textbf{Total} & \textbf{128} & \textbf{0} & \textbf{72} & \textbf{55} & \textbf{255} \\
& \textit{Percentage} & \textit{50.2\%} & \textit{0.00\%} & \textit{28.2\%} & \textit{21.6\%} & \\
& \textit{Percentage excluding DS-V3} & \textit{50.3\%} & \textit{0.0\%} & \textit{27.2\%} & \textit{22.5\%} \\
\midrule
\multicolumn{7}{l}{\textbf{ZL-XXL}} \\
& Gemini 2.5 Pro + ASP & 0 & 0 & 8 & 13 & 21 \\
& DS-R1-0528 + ASP & 23 & 7 & 14 & 9 & 53 \\
& o4-mini + ASP & 45 & 0 & 11 & 15 & 71 \\
& Gemini 2.5 Flash + ASP & 45 & 1 & 20 & 32 & 98 \\
& DS-V3 + ASP & 51 & 0 & 8 & 30 & 89 \\
\cmidrule(l){2-7}
& \textbf{Total} & \textbf{164} & \textbf{8} & \textbf{61} & \textbf{99} & \textbf{332} \\
& \textit{Percentage} & \textit{49.4\%} & \textit{2.41\%} & \textit{18.4\%} & \textit{29.82\%} & \\
& \textit{Percentage excluding DS-V3} & \textit{46.5\%} &\textit{ 3.3\% }& \textit{21.8\%} &\textit{ 28.4\%} \\
\midrule
\multicolumn{7}{l}{\textbf{SudokuBench}} \\
& Gemini 2.5 Pro + ASP & 20 & 0 & 6 & 13 & 39 \\
& DS-R1-0528 + ASP & 46 & 0 & 11 & 15 & 72 \\
& o4-mini + ASP & 67 & 1 & 4 & 7 & 79 \\
& Gemini 2.5 Flash + ASP & 48 & 0 & 6 & 11 & 65 \\
& DS-V3 + ASP & 70 & 0 & 8 & 10 & 88 \\
\cmidrule(l){2-7}
& \textbf{Total} & \textbf{251} & \textbf{1} & \textbf{35} & \textbf{56} & \textbf{343} \\
& \textit{Percentage} & \textit{73.18\%} & \textit{0.29\%} & \textit{10.20\%} & \textit{16.33\%} & \\
& \textit{Percentage excluding DS-V3} & \textit{71.0\%} & \textit{0.4\% }& \textit{10.6\%} & \textit{18.0\%} \\
\midrule[\heavyrulewidth]
\multicolumn{2}{l}{\textbf{Total}} & \textbf{543} & \textbf{9} & \textbf{168} & \textbf{210} & \textbf{930} \\
\multicolumn{2}{l}{\textbf{Total excluding DS-V3}} & \textbf{381} & \textbf{9} & \textbf{127} & \textbf{154} & \textbf{671} \\
\multicolumn{2}{l}{\textit{\textbf{Total Percentage}}} & \textit{\textbf{58.39\%}} & \textit{\textbf{0.97\%}} & \textit{\textbf{18.06\%}} & \textit{\textbf{22.58\%}} & \\
\multicolumn{2}{l}{\textit{\textbf{Total Percentage excluding DS-V3 }}} & 56.8\% & 1.3\% & 18.9\% & 23.0\%\\
\bottomrule
\end{tabular}
\end{table*}

\begin{table*}[htbp]
\centering
\caption{Error cases with revisions and without a reference.}
\label{tab:error_cases-with-revisions-without-reference}
\footnotesize 
\begin{tabular}{@{}llrrrrr@{}}
\toprule
\textbf{Dataset} & \textbf{Model} & \textbf{Syntax} & \textbf{Timeout} & \textbf{Sat} & \textbf{Unsat} & \textbf{Total} \\
\midrule
\multicolumn{7}{l}{\textbf{ZL-XL}} \\
& Gemini 2.5 Pro + ASP & 0 & 0 & 2 & 1 & 3 \\
& DS-R1-0528 + ASP & 0 & 0 & 3 & 2 & 5 \\
& o4-mini + ASP & 0 & 0 & 1 & 2 & 3 \\
& Gemini 2.5 Flash + ASP & 0 & 0 & 1 & 61 & 62 \\
& DS-V3 + ASP & 6 & 1 & 20 & 24 & 51 \\
\cmidrule(l){2-7}
& \textbf{Total} & \textbf{6} & \textbf{1} & \textbf{27} & \textbf{90} & \textbf{124} \\
& \textit{Percentage} & \textit{4.84\%} & \textit{0.81\%} & \textit{21.77\%} & \textit{72.58\%} & \\
& \textit{Percentage excluding DS-V3} & \textit{0.00\%} & \textit{0.00\%} & \textit{9.59\%} & \textit{90.41\%} & \\
\midrule
\multicolumn{7}{l}{\textbf{ZL-XXL}} \\
& Gemini 2.5 Pro + ASP & 0 & 0 & 1 & 11 & 12 \\
& DS-R1-0528 + ASP & 0 & 5 & 15 & 4 & 24 \\
& o4-mini + ASP & 1 & 2 & 15 & 15 & 33 \\
& Gemini 2.5 Flash + ASP & 0 & 0 & 6 & 69 & 75 \\
& DS-V3 + ASP & 19 & 3 & 18 & 32 & 72 \\
\cmidrule(l){2-7}
& \textbf{Total} & \textbf{20} & \textbf{10} & \textbf{55} & \textbf{131} & \textbf{216} \\
& \textit{Percentage} & \textit{9.26\%} & \textit{4.63\%} & \textit{25.46\%} & \textit{60.65\%} & \\
& \textit{Percentage excluding DS-V3} & \textit{0.69\%} & \textit{4.86\%} & \textit{25.69\%} & \textit{68.75\%} & \\
\midrule
\multicolumn{7}{l}{\textbf{SudokuBench}} \\
& Gemini 2.5 Pro + ASP & 5 & 0 & 14 & 5 & 24 \\
& DS-R1-0528 + ASP & 1 & 1 & 14 & 16 & 32 \\
& o4-mini + ASP & 2 & 0 & 15 & 11 & 28 \\
& Gemini 2.5 Flash + ASP & 0 & 0 & 10 & 34 & 44 \\ 
& DS-V3 + ASP & 33 & 1 & 20 & 24 & 78 \\
\cmidrule(l){2-7}
& \textbf{Total} & \textbf{41} & \textbf{2} & \textbf{73} & \textbf{90} & \textbf{206} \\ 
& \textit{Percentage} & \textit{19.90\%} & \textit{0.97\%} & \textit{35.44\%} & \textit{43.69\%} & \\ 
& \textit{Percentage excluding DS-V3} & \textit{6.25\%} & \textit{0.78\%} & \textit{41.41\%} & \textit{51.56\%} & \\ 
\midrule[\heavyrulewidth]
\multicolumn{2}{l}{\textbf{Total}} & \textbf{67} & \textbf{13} & \textbf{155} & \textbf{311} & \textbf{546} \\ 
\multicolumn{2}{l}{\textbf{Total excluding DS-V3}} & \textbf{9} & \textbf{8} & \textbf{97} & \textbf{231} & \textbf{345} \\
\multicolumn{2}{l}{\textit{\textbf{Total Percentage}}} & \textit{\textbf{12.27\%}} & \textit{\textbf{2.38\%}} & \textit{\textbf{28.39\%}} & \textit{\textbf{56.96\%}} & \\ 

\multicolumn{2}{l}{\textit{\textbf{Total Percentage excluding DS-V3}}} & \textit{\textbf{2.61\%}} & \textit{\textbf{2.32\%}} & \textit{\textbf{28.12\%}} & \textit{\textbf{66.96\%}} & \\
\bottomrule
\end{tabular}
\end{table*}

\section{Comparison with \citet{hao25large}'s Method} \label{app:hao}


\smallskip
\noindent
We include a preliminary comparison with the SMT-based approach \cite{hao25large}.

\citet{hao25large} demonstrate strong results using extensive task-specific prompt engineering—up to 10 handcrafted prompts with few-shot examples per domain and domain-specific schemas. 
For fair comparison, we evaluate against {\sf LLM+ASP} with no revisions and no reference, testing whether frameworks can generalize without extensive manual engineering. This methodology shift makes direct performance comparison with prior results inappropriate.
Table~\ref{tab:smt-preliminary} shows the experimental result, yielding the following observations:
\begin{table*}[t]
\centering
\footnotesize
\caption{
Zero-Shot Comparison of {\sf LLM+SMT} and {\sf LLM+ASP} (No Reference, No Revisions)}
\label{tab:smt-preliminary}
\setlength{\tabcolsep}{3.5pt}
\begin{tabular}{l|c|c|c|c|c|c|c}
\toprule
\textbf{Model} & \textbf{ZL-XL} & \textbf{ZL-XXL} & \textbf{Sudoku} & \textbf{MysteryBW} & \textbf{Skeptical} & \textbf{Credulous} & \textbf{BoardQA} \\
 & (\%) & (\%) & (\%) & (\%) & (\%) & (\%) & (\%) \\
\midrule
\multicolumn{8}{c}{\textbf{LLM + SMT (Zero-Shot, No Task-Specific Engineering)}} \\
\midrule
Gemini 2.5 Pro+SMT    & 44.0 & 35.0  & 51.0 & 47.0 & 33.0  & 19.5  & 50.0 \\
Gemini 2.5 Flash+SMT    & 41.0 & 46.0   &  38.0 & 45.0  & 16.0   & 19.5  & 40.0 \\
o4-mini+SMT    & 51.0  &  45.0  & 19.0 & 38.0 & 34.0 & 38.0 & 54.5 \\
DS-R1-0528 + SMT    & 31.0  & 23.0  & 31.0  & 37.0  &  29.0 & 33.0 & 43.5 \\
DS-V3 + SMT    & 6.0 & 3.0   & 20.0  & 2.0 & 25.0  & 26.5 & 33.0 \\
\midrule
Average & 34.6 & 31.0 & 31.8 & 33.8 & 27.4 & 27.3 & 44.2 \\
\midrule
Gemini 2.5 Pro + ASP & 88.0 & 68.0 & 71.0 & 48.0 & 98.5 & 96.0 & 67.5  \\
Gemini 2.5 Flash + ASP & 15.0 & 2.0 & 35.0 & 23.0 & 96.5 & 97.5 & 67.5  \\
o4-mini + ASP & 58.0 & 29.0 & 21.0 & 5.0 & 93.0 & 89.5 & 71.0  \\
DS-R1-0528 + ASP & 64.0 & 47.0 & 28.0 & 24.0 & 86.5 & 89.0 & 74.0  \\
DS-V3 + ASP & 18.0 & 11.0 & 12.0 & 3.0 & 68.5 & 65.0 & 37.0  \\
\midrule
Average & 48.6 & 31.4 & 33.4 & 20.6 & 88.6 & 87.4 & 63.4 \\
\bottomrule
\end{tabular}
\end{table*}

\smallskip
\noindent
{\bf (1) Fundamental advantages for nonmonotonic reasoning.}\ \ 
The performance gap is most instructive on nonmonotonic reasoning. On \textbf{MultiLogicNMR}, zero-shot {\sf LLM+SMT} achieves only 27.4\% (skeptical) and 27.3\% (credulous)—barely above random chance (33.3\%). {\sf LLM+ASP} achieves 88.6\% and 87.4\%—a more than 3$\times$ improvement (See Appendix~\ref{appendix:iterative-corrections-no-reference} for {\sf  LLM+ASP} without reference but with revisions). 

{\sf LLM+SMT} frequently outputs ``unknown'' for \textbf{MultiLogicNMR} (91\% of skeptical tasks, 70.5\% of credulous tasks, versus 33\% correct rate)—a consequence of forcing nonmonotonic reasoning into monotonic frameworks, particularly the inability to express default negation.

This reflects a \textbf{fundamental theoretical limitation}: SMT solvers operate under monotonic logic, where adding information can only derive new conclusions, never retract them. Defeasible reasoning and exception handling require sophisticated, domain-specific SMT encodings that demand expert knowledge.
On the other hand, ASP, designed explicitly for nonmonotonic reasoning via stable model semantics, treats default rules, exceptions, and alternative models as first-class constructs. LLMs can express these patterns in ASP with minimal guidance.

\smallskip
\noindent
{\bf (2) Zero-shot generalization challenges.}\ \ 
Even on tasks well-suited to SMT (constraint satisfaction, planning), zero-shot {\sf LLM+SMT} shows modest performance: \textbf{ZL-XL} (34.6\%), \textbf{ZL-XXL} (31.0\%), \textbf{SudokuBench} (31.8\%), \textbf{Mystery Blocksworld} (33.8\%)—sometimes underperforming baseline LLMs.

Without careful prompting, LLMs make persistent encoding errors (incorrect scoping, malformed constraints, missing axioms), suggesting either: (a) SMT encodings require additional task-specific guidance to yield reasonable performance, or (b) LLMs require more extensive exposure to SMT code during training.

\section{Reference Generation} \label{appendix:compact-reference}

\NBB{give prompt; give full path for the references }

We present the prompts used to generate the standard and compact reference here, along with the compact reference. We place the longer standard reference on the GitHub page.

The standard reference is generated by supplying the PDF file of the Clingo Guide (\url{https://github.com/potassco/guide/releases}) and the following prompt.

\begin{lstlisting}[escapechar=, escapeinside=, mathescape=false]
Rewrite this PDF so that LLMs can effectively understand the materials. Only include Chapters 2, 3, 6, and 11. Do not omit any details from them. For any figures verbalize them so that LLMs can understand them. Do not include any citations like "[cite: ###]". Do not use raw byte sequences like like "<0xE2><0x82><0x9B>", etc. Rather, use appropriate latex notation (like _s for subscript "s", or \rightarrow for right arrow, etc.). Encapsulate the entire text with backticks. Use 4 backticks ("````") in the beginning and end, and within this textblock can be any formatting.

\end{lstlisting}

The compact reference is generated by supplying the standard reference as a text file, along with the following short prompt.

\begin{lstlisting}[escapechar=, escapeinside=, mathescape=false]
Attached is a guide about Clingo/ASP. Please produce another one which only includes basics (e.g., rules, variables, operators, choice rules, etc.), pooling, optimization, and action planning (including incremental clingo). Do not include any citations like "[cite: ###]". Do not use raw byte sequences like like "<0xE2><0x82><0x9B>", etc. Use the same formatting. Encapsulate the entire text with backticks. Use 4 backticks ("````") in the beginning and end, and within this textblock can be any formatting.
\end{lstlisting}

The following is the compact reference:

\begin{lstlisting}[escapechar=, escapeinside=, mathescape=false]
Answer Set Programming (ASP) is a declarative approach to problem-solving. You describe the problem using logical rules, and a solver, like `clingo`, finds the solutions (called "answer sets").

### Basic Structure

An ASP program typically separates the problem **instance** from its general **encoding**.

* **Instance:** A set of simple facts that describes a specific scenario. For example, in a graph coloring problem, the facts would define the nodes and edges.
* **Encoding:** A set of rules with variables that defines the logic and constraints of the problem for any instance.

The common structure for an encoding is **Generate-Define-Test**:
1.  **Generate:** Create a space of potential solution candidates. This is often done with a **choice rule**.
2.  **Define:** Use rules to define auxiliary concepts based on the generated candidates.
3.  **Test:** Use **integrity constraints** to eliminate candidates that violate the problem's rules.

---

### Core Language Concepts

A `clingo` program consists of rules ending with a period (`.`).

* **Facts:** Unconditional statements, like `node(1).` or `edge(1,2).`.
* **Rules:** Have a head and a body, written as `head :- body.`. This means "if the body is true, the head must be true." The body is a comma-separated list of conditions.
* **Integrity Constraints:** Rules with no head, written as `:- body.`. This forbids solutions where the body is true.

**Variables and Terms**
* **Variables** start with an uppercase letter (e.g., `X`, `Node`).
* **Constants** (or symbols) start with a lowercase letter (e.g., `a`, `peg1`).
* Rules must be **safe**: every variable in a rule must appear in a positive literal in the body, which ensures it gets a concrete value during solving.

**Choice Rules and Aggregates**
The core of the "Generate" step is the choice rule, which uses curly braces `{}`.

* `{ color(X,C) : color(C) } = 1 :- node(X).`
    This rule means: "For each node `X`, choose exactly one color `C` for it". The `{...}` part is a cardinality constraint.

Aggregates like `#count`, `#sum`, `#min`, and `#max` can be used in the body to test properties of a solution.
* `:- #count { student(S) : enrolled(S) } < 3.`
    This constraint says a solution is invalid if fewer than 3 enrolled students exist.

---

### Operators, Pooling, and Intervals

`clingo` supports built-in operators that are evaluated by the grounder.

* **Operators:**
    * `+` (Addition)
    * `-` (Subtraction, Unary Minus)
    * `*` (Multiplication)
    * `/` (Integer Division)
    * `\` (Modulo)
    * `**` (Exponentiation)
    * `|...|` (Absolute Value, e.g., `|X-Y|`)
    * `&` (Bitwise AND)
    * `?` (Bitwise OR)
    * `^` (Bitwise XOR)
    * `~` (Bitwise Complement)

* **Comparison Operators:** Terms can be compared with `=`, `!=`, `<`, `>`, `<=`, and `>=`. These are used as conditions in the body of rules.
    * `eligible(P) :- person(P), age(P,A), A >= 18.`

* **Pooling:** The semicolon (`;`) is syntactic sugar for defining multiple facts or rules concisely.
    * **Syntax:** `f(arg1, (X;Y), arg3)` is shorthand for `f(arg1, X, arg3)` and `f(arg1, Y, arg3)`. (note: the parentheses are needed).
    * **Expansion:** Similar to intervals.
        * **In Head:** Conjunctive expansion. `p(a;b).` expands to `p(a). p(b).` `p((a;b), (1;2)).` expands to `p(a,1). p(a,2). p(b,1). p(b,2).`
        * **In Body (typically with `=`):** Disjunctive expansion. `p(X) :- X = (a;b).` is equivalent to rules generating `p(a).` and `p(b).`.

* **Example:**
```prolog
peg(a;b;c). % Expands to peg(a). peg(b). peg(c).
test((thing), (a;b;c;d)). % Expands to test(thing,a). test(thing,b). test(thing,c). test(thing,d).
```

* **Intervals:** A range of integers can be defined with `..`.
    * `disk(1..4).` is shorthand for `disk(1). disk(2). disk(3). disk(4).`

---

### Optimization

To find the best solution, you can use optimization statements.
**Example:** Minimize the total cost of the selected cycle edges.
* `#minimize { Cost,X,Y : cycle(X,Y), cost(X,Y,Cost) }.`
    This statement tells the solver to find a solution where the sum of `Cost` for all selected `cycle(X,Y)` atoms is as small as possible.

**Explanation:**
* The `Generate` part uses cardinality constraints to ensure every node has exactly one incoming and one outgoing edge selected in the `cycle/2` predicate.
* The `Define` and `Test` parts work together to ensure these selected edges form a single cycle that includes all nodes (a Hamiltonian cycle). The `reached/1` predicate, starting from node 1, uses recursion. ASP's stable model semantics ensure that only atoms derivable from non-circular paths are included. The integrity constraint `: - node(Y), not reached(Y).` eliminates answer sets where the selected edges form multiple disjoint cycles or do not cover all nodes.
* The `Optimize` part uses `#minimize` to sum the costs of the edges selected in `cycle/2` and finds the answer set(s) where this sum is minimal. The `Cost,X,Y` part in the minimize statement ensures that each distinct edge `(X,Y)` with its specific cost `Cost` contributes to the sum (handling potential multiple edges between nodes if the instance allowed them, though not strictly necessary here).

* `#maximize` works similarly to find the largest possible value.

### **Blocks World Planning**

**Problem:** Find a sequence of actions (plan) to transform an initial configuration of blocks on a table into a desired goal configuration. We want the *shortest* plan (minimum number of actions).

**Challenge:** Finding the shortest plan often requires trying plans of length 1, then 2, then 3, etc., until a solution is found. clingo supports an *incremental* mode to handle this efficiently, avoiding redundant grounding and solving at each step.

#### **6.3.1 Problem Instance**

Facts define blocks, the initial state, and the goal state. on(Block, Location) describes the state, where Location can be another block or table.

**Instance:**

% Blocks  
block(b0). block(b1). block(b2).

% Initial State:  
%   b2  
%   b0  b1  
% \--------- (table)  
init(on(b1,table)).  
init(on(b2,b0)).  
init(on(b0,table)).

% Goal State:  
%   b2  
%   b1  
%   b0  
% \--------- (table)  
goal(on(b1,b0)).  
goal(on(b2,b1)).  
goal(on(b0,table)).

*(Note: Compound terms like on(b1, table) are used as arguments).*

#### **6.3.2 Problem Encoding (Incremental)**

This encoding uses \#program directives for incremental solving. clingo automatically handles grounding step-by-step.

% Include clingo's incremental mode definitions  
\#include \<incmode\>.

% \--- Static Part (Time 0\) \---  
\#program base.  
% Define possible locations  
location(table).  
location(X) :- block(X). % Blocks are also locations.  
% Define initial state using 'holds(Fact, TimeStep)'  
holds(F,0) :- init(F). % F will be instantiated with terms like on(b1,table).

% \--- Dynamic Part (Transitions for step T \> 0\) \---  
\#program step(t). % 't' is the time step parameter  
% Generate: Exactly one move action per time step.  
% move(BlockToMove, TargetLocation, TimeStep)  
{ move(X,Y,t) : block(X), location(Y), X \!= Y } \= 1\.

% Test: Constraints on moves.  
% Cannot move X if something (A) is on top of it.  
:- move(X,Y,t), holds(on(A,X),t-1).  
% Cannot move X onto Y if Y is a block and already occupied by B (unless B is X itself \- allows void moves technically, though less useful here).  
:- move(X,Y,t), holds(on(B,Y),t-1), B \!= X, Y \!= table.

% Define: Effects of actions and inertia.  
% moved(X,t): True if block X was moved at step t.  
moved(X,t) :- move(X,Y,t).  
% Direct effect: If X is moved to Y at t, then on(X,Y) holds at t.  
holds(on(X,Y),t) :- move(X,Y,t).  
% Inertia: If on(X,Z) held at t-1 and X was not moved at t, it still holds at t.  
holds(on(X,Z),t) :- holds(on(X,Z),t-1), not moved(X,t).

% \--- Check Part (Applied at each step T \>= 0\) \---  
\#program check(t). % 't' is the time step parameter  
% Test: Check if the goal is met at the current step 't'.  
% query(t) is a special atom provided by incmode, true only for the current step.  
:- query(t), goal(F), not holds(F,t). % If it's query time, goal F must hold.

% Display: Only show the move actions in the final plan.  
% This \#show is global, not tied to a specific program part.  
\#show move/3.

**Explanation (incmode):**

* \#include \<incmode\>.: Enables incremental mode.  
* \#program base.: Grounded once at the beginning (step 0).  
* \#program step(t).: Grounded incrementally for steps t=1,2,3,dots. Defines transitions from step t-1 to t.  
* \#program check(t).: Grounded incrementally for steps t=0,1,2,dots. Checks if the goal is met at step t.  
* query(t): Atom automatically added by clingo in incremental mode. It is true only for the maximum step t currently being considered in a given iteration. The goal check : \- query(t), goal(F), not holds(F,t). ensures that if clingo is checking for a solution at step t, the goal must be satisfied at that step.  
* **Requirement:** For correct incremental grounding, atoms defined in step(t) (like holds(F,t)) must be distinct from atoms defined in base or step(t') for t'neqt. Using the time step t as an argument ensures this.
\end{lstlisting}

\end{document}